\documentclass{article} 
\usepackage{iclr2024_conference,times}

\usepackage{amsmath,amsfonts,bm}

\def\eqref#1{equation~\ref{#1}}

\def\1{\bm{1}}

\DeclareMathAlphabet{\mathsfit}{\encodingdefault}{\sfdefault}{m}{sl}
\SetMathAlphabet{\mathsfit}{bold}{\encodingdefault}{\sfdefault}{bx}{n}

\usepackage{hyperref}
\usepackage{url}
\usepackage{booktabs}       
\usepackage{amsfonts}       
\usepackage{amsmath, amssymb}
\usepackage{nicefrac}       
\usepackage{xcolor}         
\usepackage{subcaption}
\usepackage{graphicx}
\usepackage{multirow}
\usepackage{colortbl}
\usepackage{wrapfig}
\usepackage{listings}
\definecolor{Gray}{gray}{0.92}
\newcolumntype{g}{>{\columncolor{Gray}}c}

\graphicspath{ {images/} {images-suppl/} }

\definecolor{codegreen}{rgb}{0,0.6,0}
\definecolor{codegray}{rgb}{0.5,0.5,0.5}
\definecolor{codepurple}{rgb}{0.58,0,0.82}
\definecolor{backcolour}{rgb}{1,1,1}

\lstdefinestyle{mystyle}{
    backgroundcolor=\color{backcolour},   
    commentstyle=\color{codegreen},
    keywordstyle=\color{magenta},
    numberstyle=\tiny\color{codegray},
    stringstyle=\color{codepurple},
    basicstyle=\ttfamily\footnotesize,
    breakatwhitespace=false,         
    breaklines=true,                 
    captionpos=b,                    
    keepspaces=true,                 
    numbers=left,                    
    numbersep=5pt,                  
    showspaces=false,                
    showstringspaces=false,
    showtabs=false,                  
    tabsize=2
}

\title{FedWon: Triumphing Multi-domain Federated Learning Without Normalization}

\author{Weiming Zhuang \\
Sony AI \\
\texttt{weiming.zhuang@sony.com} \\
\And
Lingjuan Lyu \\
Sony AI \\
\texttt{lingjuan.lv@sony.com} \\
}

\iclrfinalcopy 
\begin{document}

\maketitle

\begin{abstract}
   Federated learning (FL) enhances data privacy with collaborative in-situ training on decentralized clients. Nevertheless, FL encounters challenges due to non-independent and identically distributed (non-i.i.d) data, leading to potential performance degradation and hindered convergence. While prior studies predominantly addressed the issue of skewed label distribution, our research addresses a crucial yet frequently overlooked problem known as multi-domain FL. In this scenario, clients' data originate from diverse domains with distinct feature distributions, instead of label distributions. To address the multi-domain problem in FL, we propose a novel method called \textbf{Fed}erated learning \textbf{W}ith\textbf{o}ut \textbf{n}ormalizations (FedWon). 
   FedWon draws inspiration from the observation that batch normalization (BN) faces challenges in effectively modeling the statistics of multiple domains, while existing normalization techniques possess their own limitations. In order to address these issues, FedWon eliminates the normalization layers in FL and reparameterizes convolution layers with scaled weight standardization. 
   Through extensive experimentation on five datasets and five models, our comprehensive experimental results demonstrate that FedWon surpasses both FedAvg and the current state-of-the-art method (FedBN) across all experimental setups, achieving notable accuracy improvements of more than 10\% in certain domains. Furthermore, FedWon is versatile for both cross-silo and cross-device FL, exhibiting robust domain generalization capability, showcasing strong performance even with a batch size as small as 1, thereby catering to resource-constrained devices. Additionally, FedWon can also effectively tackle the challenge of skewed label distribution.
\end{abstract}

\section{Introduction}

Federated learning (FL) has emerged as a promising method for distributed machine learning, enabling in-situ model training on decentralized client data. It has been widely adopted in diverse applications, including healthcare \citep{li2019brain-tumor1,bernecker2022fednorm}, mobile devices \citep{hard2018gboard,paulik2021apple}, and autonomous vehicles \citep{zhang2021auto,nguyen2021auto1,posner2021vehicular}. However, FL commonly suffers from statistical heterogeneity, where the data distributions across clients are non-independent and identically distributed (non-i.i.d) \citep{Li2020FedChallenges}. This is due to the fact that data generated from different clients is highly likely to have different data distributions, which can cause performance degradation \citep{zhao2018non-iid,hsieh2020non,tan2023heterogeneity} even divergence in training \citep{zhuang2020fedreid,zhuang2022fedreid,wang2023batch}.

\begin{figure}[t!]
  \centering
  \begin{subfigure}[t]{0.7\textwidth}
    \includegraphics[width=\textwidth]{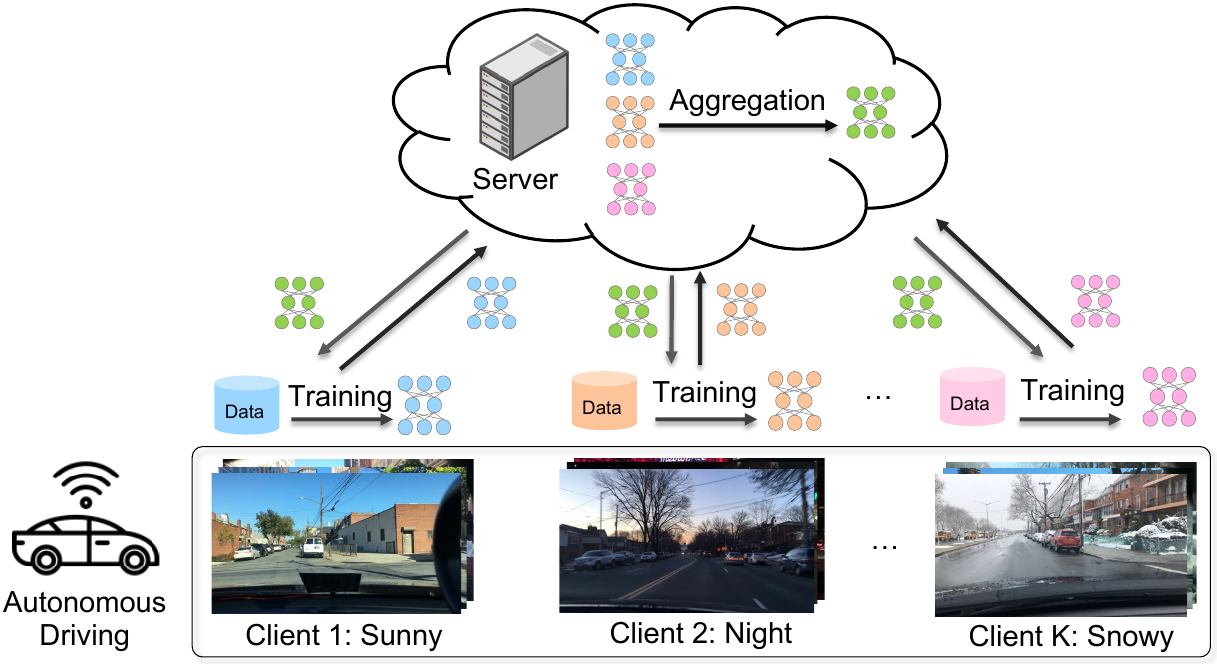}
    \caption{Illustration of Multi-domain FL.}
    \label{fig:multi-domain}
  \end{subfigure}  
  \hfill
  \begin{subfigure}[t]{0.25\textwidth}
      \includegraphics[width=\textwidth]{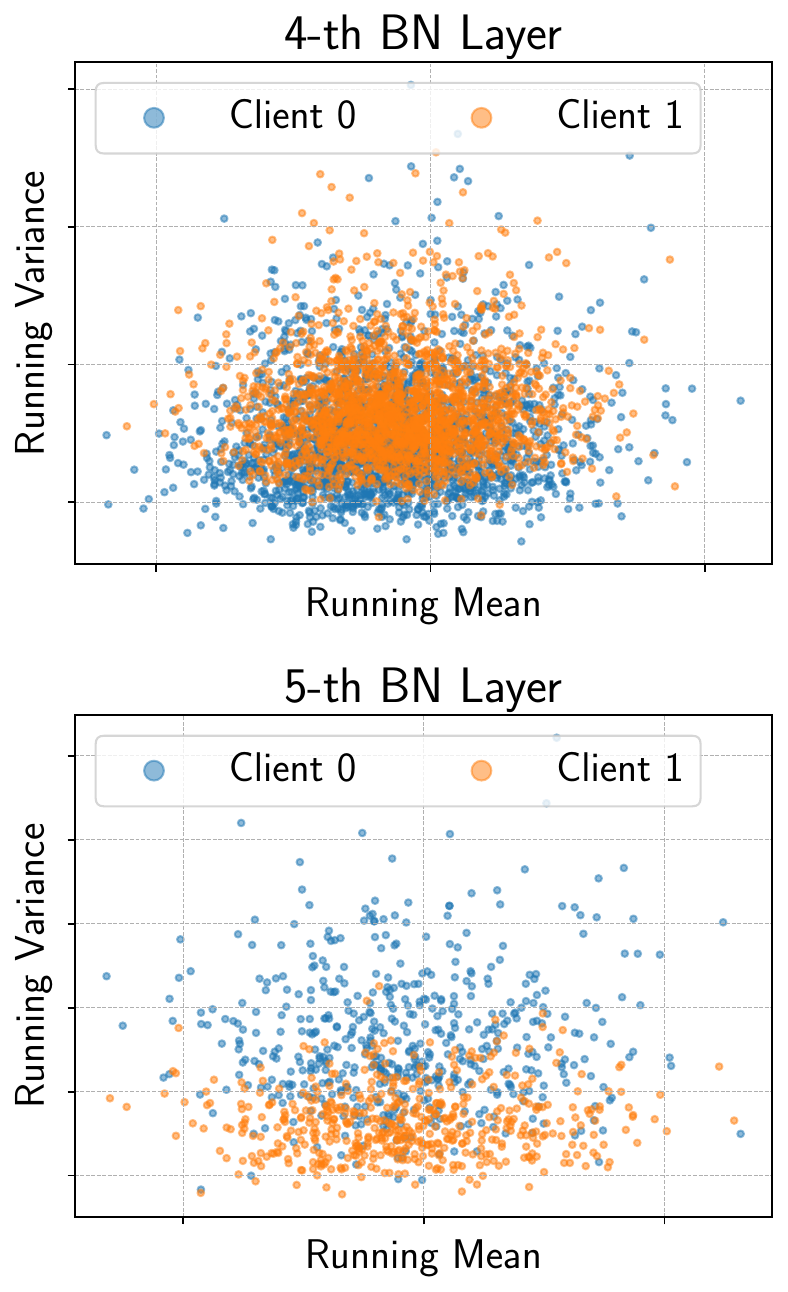}
      \caption{Statistics of BN Layers.}
      \label{fig:bn-stats}
  \end{subfigure}
  \hfill
 \caption{(a) We consider multi-domain federated learning, where each client contains data of one domain. This setting is highly practical and applicable in real-world scenarios. For example, autonomous cars in distinct locations capture images in varying weather conditions. (b) Visualization of batch normalization (BN) channel-wise statistics from two clients, each with data from a single domain. The upper and lower figures are results from the 4-th and 5-th BN layers of a 6-layer CNN, respectively. It highlights different feature statistics of BN layers trained on different domains.}
 \label{fig:overall}
\end{figure}

The majority of studies that address the problem of non-i.i.d data focus on the issue of skewed label distribution, where clients have different label distributions \citep{fedprox,hsieh2020non,wang2020fedma,chen2022calfat}. However, multi-domain FL, where clients' data are from different domains, has received less attention, despite its practicality in reality. Figure \ref{fig:multi-domain} depicts two practical examples of multi-domain FL. For example, multiple autonomous cars may collaborate on model training, but their data could originate from different weather conditions or times of day, leading to domain gaps in collected images \citep{cordts2016cityscapes,yu2020bdd100k}. Similarly, multiple healthcare institutions collaborating on medical imaging analysis may face significant domain gaps due to variations in imaging machines and protocols \citep{bernecker2022fednorm}. Developing effective solutions for multi-domain FL is a critical research problem with broad implications.

However, the existing solutions are unable to adequately address the problem of multi-domain FL. FedBN \citep{li2021fedbn} attempts to solve this problem by keeping batch normalization (BN) parameters and statistics \citep{ioffe2015bn} locally in the client, but it is only suitable for cross-silo FL \citep{kairouz2021advances}, where clients are organizations like healthcare institutions, because it requires clients to be stateful \citep{karimireddy2020scaffold} (keeping states of BN information) and participate training in every round. As a result, FedBN is not suitable for cross-device FL, where the clients are stateless and only a fraction of clients participate in training.
Besides, BN relies on the assumption that training data are from the same distribution, ensuring the mean and variance of each mini-batch are representative of the entire data distribution \citep{ioffe2015bn}. Figure \ref{fig:bn-stats} shows that the running means and variances of BNs differ significantly between two FL clients from different domains, as well as between the server and clients (statistics of all BN layers are in Figure \ref{fig:bn-stats-all} in Appendix).
Alternative normalizations like Layer Norm \citep{ba2016layer} and Group Norm \citep{wu2018group} have not been studied for multi-domain FL, 
but they have limitations like requiring extra computation in inference.

This paper explores a fundamentally different approach to address multi-domain FL. Given that BN struggles to capture multi-domain data and other normalizations come with their own limitations, we further ask the question: is normalization indispensable to learning a general global model for multi-domain FL? In recent studies, normalization-free ResNets \citep{brock2021characterizing} demonstrates comparable performance to standard ResNets\citep{he2016resnet}. Inspired by these findings, we build upon this methodology and explore its untapped potential within the realm of multi-domain FL.

We introduce \textbf{Fed}erated learning \textbf{W}ith\textbf{o}ut \textbf{n}ormalizations (FedWon) to address the domain discrepancies among clients in multi-domain FL. FedWon follows FedAvg \citep{fedavg} protocols for server aggregation and client training. Unlike existing methods, FedWon removes normalization layers (e.g., BN layers), and reparameterizes convolution layers with Scaled Weight Standardization \citep{brock2021characterizing}. We conduct extensive experiments on five datasets using five models.
The experimental results indicate that FedWon outperforms state-of-the-art methods on all datasets and models. The \textit{general global model} trained by FedWon can achieve more than 10\% improvement on certain domains compared to the \textit{personalized models} from FedBN \citep{li2021fedbn}. Moreover, our empirical evaluation demonstrated three key benefits of FedWon: 1) FedWon is versatile to support both cross-silo and cross-device FL; 2) FedWon achieves competitive performance on small batch sizes (even on a batch size of 1), which is particularly useful for resource-constrained devices; 3) FedWon can also be applied to address the skewed label distribution problem.

In summary, our contributions are as follows:

\begin{itemize}
  \item We introduce FedWon, a simple yet effective method for multi-domain FL. By removing all normalization layers and using scaled weight standardization, FedWon is able to learn a general global model from clients with significant domain discrepancies.
  \item To the best of our knowledge, FedWon is the first method that enables both cross-silo and cross-device FL without relying on any form of normalization. Our study also reveals the unexplored benefits of this method, particularly in the context of multi-domain FL.
  \item Extensive experiments demonstrate that FedWon outperforms state-of-the-art methods on all the evaluated datasets and models, and is suitable for training with small batch sizes, which is especially beneficial for cross-device FL in practice.
\end{itemize}

\begin{figure*}[t!]
    \centering
    \begin{subfigure}[t]{0.3\textwidth}
        \includegraphics[width=\textwidth]{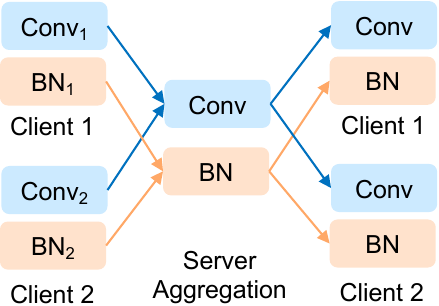}
        \caption{FedAvg}
        \label{fig:fedavg}
    \end{subfigure}
    \hfill
    \begin{subfigure}[t]{0.3\textwidth}
       \includegraphics[width=\textwidth]{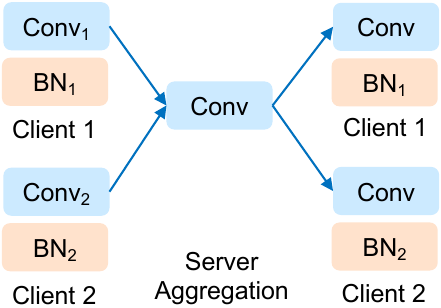}
       \caption{FedBN}
       \label{fig:fedbn}
   \end{subfigure}
   \hfill
   \begin{subfigure}[t]{0.3\textwidth}
    \includegraphics[width=\textwidth]{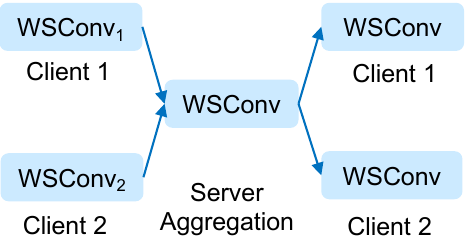}
    \caption{FedWon (Ours)}
    \label{fig:fedwon}
  \end{subfigure}
  \caption{Illustration of three FL algorithms: (a) FedAvg aggregates both convolution (Conv) layers and batch normalization (BN) layers in the server; (b) FedBN keeps BN layers in clients and only aggregates Conv layers; (c) Our proposed \textbf{Fed}erated learning \textbf{W}ith\textbf{o}ut \textbf{n}ormalizations (FedWon) removes all BN layers and reparameterizes Conv layers with scaled weight standardization (WSConv).}
  \label{fig:algorithms} 
  \end{figure*}

\section{Preliminary}

Before diving into the benefits brought by removing normalizations, we first introduce FL with batch normalization. Then, we review alternative normalization methods and normalization-free networks.

\subsection{Federated Learning with Batch Normalization}
\label{sec:fl-w-bn}

Batch normalization (BN) \citep{ioffe2015bn}, commonly used as a normalization layer, has been a fundamental component in deep neural networks (DNN). The BN operation is defined as:
\begin{equation}
  BN(x) = \gamma \frac{x - \mu}{\sqrt{\sigma^2 + \epsilon}} + \beta,
  \label{eq:bn}
\end{equation}
where mean \(\mu\) and variance \(\sigma\) are computed from a mini-batch of data, and \(\gamma\) and \(\beta\) are two learnable parameters. The term \(\epsilon\) is a small positive value that is added for numerical stability.

BN offers several benefits, including reducing internal covariate shift, stabilizing training, and accelerating convergence \citep{ioffe2015bn}. Moreover, it is more robust to hyperparameters \citep{bjorck2018understandingbn} and has smoother optimization landscapes \citep{santurkar2018doesbn}. However, its effectiveness is based on the assumption that the training data is from the same domain, such that the mean $\mu$ and variance $\sigma$ computed from a mini-batch are representative of the training data \citep{ioffe2015bn}. In centralized training, BN has been found to struggle with modeling the statistics from multiple domains, leading to the development of domain-specific BN techniques \citep{li2016revisiting,chang2019domain}. Similarly, in multi-domain FL, DNN with BN can encounter difficulties in capturing the statistics of multiple domains while training a single global model.

Federated learning (FL) trains machine learning models collaboratively from decentralized clients, coordinated by a central server \citep{kairouz2021advances}. It enhances data privacy by keeping the raw data locally on clients. FedAvg \citep{fedavg} is the most popular FL algorithm. A common issue in FL is non-i.i.d data across clients, which could lead to performance degradation and difficulties in convergence \citep{hsieh2020non,zhuang2021fedu,zhuang2022divergence,wang2023batch}. Skewed label distribution, where clients have different label distributions, is a widely discussed non-i.i.d. problem with numerous proposed solutions \citep{fedprox,zhang2023addressing,chen2022calfat}. To address this problem, multiple works provide solutions that introduce special operations on BN to personalize a model for each client \citep{lu2022personalized}. For example, SiloBN \citep{andreux2020siloed} keeps BN statistics locally in clients. FixBN \citep{zhong2023making} only trains BN statistics in the first stage of training and freezes them thereafter. FedTAN \citep{wang2023batch} tailors for BN by performing iterative layer-wise aggregations, introducing numerous extra communication rounds.

In contrast, multi-domain FL, where the data domains differ across clients, has received less attention \citep{chen2018closing,shen2022cd2}. FedBN \citep{li2021fedbn} and FedNorm \citep{bernecker2022fednorm} addresses this issue by locally keeping the BN layers in clients and aggregating only the other model parameters.  
PartialFed \citep{sun2021partialfed} keeps model initialization strategies in clients and use them to load models in new training rounds.
While these methods excel in cross-silo FL, where clients are stable and can retain statefulness, they are unsuitable for cross-device FL by design. In the latter scenario, clients are stateless, and only a fraction of clients participate in each round of training \citep{kairouz2021advances}. Besides, FMTDA \citep{yao2022federated} adapts source domain data in the server to target domains in clients, whereas we do not assume availablility of data in the server.

\subsection{Alternative Normalization Methods}
\label{sec:norm-methods}

BN has shown to be effective in modern DNNs \citep{ioffe2015bn}, but it also has limitations in various scenarios. For example, BN struggles to model statistics of training data from multiple domains \citep{li2016revisiting,chang2019domain}, and it may not be suitable for small batch sizes \citep{ioffe2017batch,wu2018group}. Researchers have proposed alternative normalizations such as Group Norm \citep{wu2018group} and Layer Norm \citep{ba2016layer}. Although these methods remove some of the constraints of BN, they come with their own 
limitations. For example, they require additional computation during inference, making them less practical for edge deployment.

Recent studies have shown that BN may not work well in FL under non-i.i.d data \citep{hsieh2020non}, due to external covariate shift \citep{du2022rethinking} and mismatch between local and global statistics \citep{wang2023batch}. Instead, researchers have adopted alternative normalizations such as GN \citep{hsieh2020non,casella2023experimenting} or LN \citep{du2022rethinking,casella2023experimenting} to mitigate the problem. 
However, these methods inherit the limitations of GN and LN in centralized training, and the recent study by \cite{zhong2023making} shows that BN and GN have no consistent winner in FL.

\subsection{Normalization-free Networks} 
\label{sec:nf-networks}

Several attempts have been made to remove normalization from DNNs in centralized training using weight initialization methods \citep{hanin2018start,zhang2019fixup,de2020batch}. Recently, \cite{brock2021characterizing} proposed a normalization-free network by analyzing the signal propagation through the forward pass of the network. Normalization-free network stabilizes training with scaled weight standardization that reparameterizes the convolution layer to prevent the mean shift in the hidden activations \citep{brock2021characterizing}. This approach achieves competitive performance compared to networks with BN on ResNet \citep{he2016resnet} and EfficientNet \citep{tan2019efficientnet}. Building on this work, Brock et al. further introduced an adaptive gradient clipping (AGC) method that enables training normalization-free networks with large batch sizes \citep{brock2021high}.

\section{Federated Learning Without Normalization}

In this section, we present the problem setup of multi-domain FL and propose FL without normalization to address the problem of multi-domain FL.

\subsection{Problem Setup}

The standard federated learning aims to train a model with parameters \(\theta\) collaboratively from total $N \in \mathbb{N}$ decentralized clients. The goal is to optimize the following problem:
\begin{equation}
  \min_{\theta \in \mathbb{R}^d} f(\theta) := \sum_{k=1}^K p_k f_k(\theta) := \sum_{k=1}^K p_k \mathbb{E}_{\xi_k \sim \mathcal{D}_k}[f_k(\theta;\xi_k)],
  \label{eq:fl}
\end{equation}
where $K \in \mathbb{N}$ is the number of participated clients (\( K \le N \)), $f_k(\theta)$ is the loss function of client $k$, $p_k$ is the weight for model aggregation in the server, and $\xi_k$ is the data sampled from distribution $\mathcal{D}_k$ of client $k$. FedAvg \citep{fedavg} sets $p_k$ to be proportional to the data size of client $k$. Each client trains for $E \in \mathbb{N}$ local epochs before communicating with the server.

Assume there are $N$ clients in FL and each client $k$ contains $n_k \in \mathbb{N}$ data samples \( \{(x_i^k, y_i^k)\}_{i=1}^{n_k} \). Skewed label distribution refers to the scenario where data in clients have different label distributions, i.e. the marginal distributions $\mathcal{P}_k(y)$ may differ across clients ($\mathcal{P}_k(y) \nsim \mathcal{P}_{k'}(y)$ for different clients $k$ and $k'$). In contrast, this work focuses on multi-domain FL, where clients possess data from various domains, and data samples within a client belong to the same domain \citep{kairouz2021advances,li2021fedbn}. 
Specifically, the marginal distribution $\mathcal{P}_k(x)$ may vary across clients ($\mathcal{P}_k(x) \nsim \mathcal{P}_{k'}(x)$ for different clients $k$ and $k'$). 
Within each client, the data samples, represented as $x_i$ and $x_j$, drawn from the same marginal distribution $\mathcal{P}_k(x)$ holds that
$\mathcal{P}_k(x_i)\sim \mathcal{P}_k(x_j)$ for all $i,j \in {1,2,...,n_k}$.
Figure \ref{fig:multi-domain} illustrates practical examples of multi-domain FL. For example, autonomous cars in different locations could capture images under different weather conditions.

\subsection{Normalization-Free Federated Learning}

Figure \ref{fig:bn-stats} demonstrates that the BN statistics of clients with data from distinct domains are considerably dissimilar in multi-domain FL. Although various existing approaches have attempted to address this challenge by manipulating or replacing the BN layers with other normalization layers \citep{li2021fedbn,du2022rethinking,zhong2023making}, they come with their own set of limitations, such as additional computation cost during inference. 
To bridge this gap, we propose a novel approach called \textbf{Fed}erated learning \textbf{W}ith\textbf{o}ut \textbf{n}ormalizations (FedWon)that removes all normalization layers in FL.

\begin{wrapfigure}{r}{3.3cm}
   \vspace{-0.4cm}
   \centering
   \includegraphics[width=0.23\textwidth]{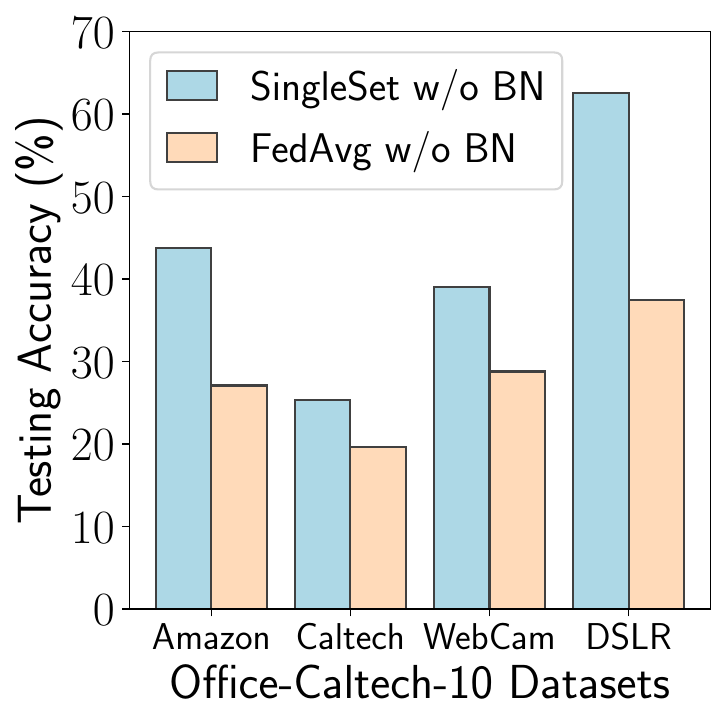}
   \caption{FedAvg without (w/o) BN yields inferior results.}
   \label{fig:no-bn}
   \vspace{-0.4cm}
\end{wrapfigure}

However, simply removing all normalization layers would lead to deteriorated performance in FL. Figure \ref{fig:no-bn} compares the performance of training in a single dataset (SingleSet) and FedAvg without normalization on four domains of the Office-Caltech-10 dataset (Further details on the experimental setup are provided in Section \ref{sec:experiments}). FedAvg without (w/o) BN yields inferior results compared to SingleSet w/o BN. The domain gaps among clients could amplify the challenges in FL when training without BNs.

Compared with FedAvg \citep{fedavg}, our proposed FedWon completely removes the normalization layers in DNNs and further reparameterizes the convolutions layer. We employ the Scaled Weight Standardization technique proposed by \cite{brock2021characterizing} to reparameterize the convolution layers after removing BN. The reparameterization formula can be expressed as follows:
\begin{equation}
  \hat{W}_{i,j} = \gamma \frac{W_{i,j} - \mu_i}{\sigma_i \sqrt{N}},
  \label{eq:wsconv}
\end{equation}
where $W_{i,j}$ is the weight matrix of a convolution layer with $i$ as the output channel and $j$ as the input channel, $\gamma$ is a constant number, $N$ is the fan-in of convolution layer,  $\mu_i = (1/N) \sum_j W_{i,j}$ and $\sigma_i^2 = (1/N) \sum_j (W_{i,j} - \mu_i)$ are the mean and variance of the $i$-th row of $W_{i,j}$, respectively. 
By removing normalization layers, FedWon eliminates batch dependency, resolves discrepancies between training and inference, and does not require computation for normalization statistics in inference. We term this parameterized convolution as WSConv.

Figure \ref{fig:algorithms} highlights the algorithmic differences between our proposed FedWon and the other two FL algorithms: 
FedAvg \citep{fedavg} and FedBN \citep{li2021fedbn}. FedAvg aggregates both convolution and BN layers on the server; FedBN only aggregates the convolution layers and keeps BN layers locally in clients. Unlike these two methods, FedWon removes BN layers, replaces convolution layers with WSConv, and only aggregates these reparameterized convolution layers. Prior work theoretically shows that BN slows down and biases the FL convergence \citep{wang2023batch}. FedWon circumvents these issues by removing BN while preserving the convergence speed that BN typically facilitates. Furthermore, FedWon offers unexplored benefits to multi-domain FL, including versatility for both cross-silo and cross-device FL, enhanced domain generalization, and compelling performance on small batch sizes, including a batch size as small as 1.

\begingroup
\setlength{\tabcolsep}{0.32em}
\begin{table}[t]\centering
  \caption{Testing accuracy (\%) comparison of different methods on three datasets. Our proposed FedWon outperforms existing methods in most of the domains. FedWon achieves the best average testing accuracy in all datasets.}\label{tab:cross-silo-mean}
  \begin{tabular}{l|l|cccccccc|c}\toprule
  &Domains &SingleSet &FedAvg &FedProx &+GN $^a$ &+LN $^b$ &SiloBN &FixBN &FedBN &Ours \\\midrule
  \multirow{6}{*}{\rotatebox[origin=c]{90}{Digit-Five}} &MNIST &94.4 &96.2 &96.4 &96.4 &96.4 &96.2 &96.3 &96.5 &\textbf{96.8} \\
  & SVHN &67.1 &71.6 &71.0 &76.9 &75.2 &71.3 &71.3 &77.3 &\textbf{77.4} \\
  & USPS &95.4 &96.3 &96.1 &96.6 &96.4 &96.0 &96.1 &96.9 &\textbf{97.0} \\
  & SynthDigits &80.3 &86.0 &85.9 &86.6 &85.6 &86.0 &85.8 &86.8 &\textbf{87.6} \\
  & MNIST-M &77.0 &82.5 &83.1 &83.7 &82.2 &83.1 &83.0 &\textbf{84.6} &84.0 \\
  &\cellcolor[HTML]{efefef}Average&\cellcolor[HTML]{efefef}83.1 &\cellcolor[HTML]{efefef}86.5 &\cellcolor[HTML]{efefef}86.5 &\cellcolor[HTML]{efefef}88.0 &\cellcolor[HTML]{efefef}87.1 &\cellcolor[HTML]{efefef}86.5 &\cellcolor[HTML]{efefef}86.5 &\cellcolor[HTML]{efefef}88.4 &\cellcolor[HTML]{efefef}\textbf{88.5} \\\midrule
  \multirow{5}{*}{\rotatebox[origin=c]{90}{Caltech-10}} & Amazon &54.5 &61.8 &59.9 &60.8 &55.0 &60.8 &59.2 &\textbf{67.2} &67.0 \\
  & Caltech &40.2 &44.9 &44.0 &50.8 &41.3 &44.4 &44.0 &45.3 &\textbf{50.4} \\
  & DSLR &81.3 &77.1 &76.0 &88.5 &79.2 &76.0 &79.2 &85.4 &\textbf{95.3} \\
  & Webcam &89.3 &81.4 &80.8 &83.6 &71.8 &81.9 &79.6 &87.5 &\textbf{90.7} \\
  &\cellcolor[HTML]{efefef}Average&\cellcolor[HTML]{efefef}66.3 &\cellcolor[HTML]{efefef}66.3 &\cellcolor[HTML]{efefef}65.2 &\cellcolor[HTML]{efefef}70.9 &\cellcolor[HTML]{efefef}61.8 &\cellcolor[HTML]{efefef}65.8 &\cellcolor[HTML]{efefef}65.5 &\cellcolor[HTML]{efefef}71.4 &\cellcolor[HTML]{efefef}\textbf{75.6} \\\midrule
  \multirow{7}{*}{\rotatebox[origin=c]{90}{DomainNet}} & Clipart &42.7 &48.9 &51.1 &45.4 &42.7 &51.8 &49.2 &49.9 &\textbf{57.2} \\
  & Infograph &24.0 &26.5 &24.1 &21.1 &23.6 &25.0 &24.5 &28.1 &\textbf{28.1} \\
  & Painting &34.2 &37.7 &37.3 &35.4 &35.3 &36.4 &38.2 &40.4 &\textbf{43.7} \\
  & Quickdraw &\textbf{71.6} &44.5 &46.1 &57.2 &46.0 &45.9 &46.3 &69.0 &69.2 \\
  & Real &51.2 &46.8 &45.5 &50.7 &43.9 &47.7 &46.2 &55.2 &\textbf{56.5} \\
  & Sketch &33.5 &35.7 &37.5 &36.5 &28.9 &38.0 &37.4 &38.2 &\textbf{51.9} \\
  &\cellcolor[HTML]{efefef}Average &\cellcolor[HTML]{efefef}42.9 &\cellcolor[HTML]{efefef}40.0 &\cellcolor[HTML]{efefef}40.2 &\cellcolor[HTML]{efefef}41.1 &\cellcolor[HTML]{efefef}36.7 &\cellcolor[HTML]{efefef}40.8 &\cellcolor[HTML]{efefef}40.3 &\cellcolor[HTML]{efefef}46.8 &\cellcolor[HTML]{efefef}\textbf{51.1} \\
  \bottomrule
  \end{tabular}
  \footnotesize{$^a$+GN means FedAvg+GN, $^b$+LN means FedAvg+LN}
\end{table}
\endgroup

\section{Experiments on Multi-domain FL}
\label{sec:experiments}

In this section, we start by introducing the experimental setup for multi-domain FL. We then validate that FedWon outperforms existing methods in both cross-silo and cross-device FL and achieves comparable performance even with a batch size of 1.
We end by providing ablation studies.

\subsection{Experiment Setup}

\paragraph{Datasets.} We conduct experiments for multi-domain FL using three datasets: Digits-Five \citep{li2021fedbn}, Office-Caltech-10 \citep{gong2012office-caltech10}, and DomainNet \citep{peng2019domainnet}. Digits-Five consists of five sets of 28x28 digit images, including MNIST \citep{lecun1998mnist}, SVHN \citep{netzer2011svhn}, USPS \citep{hull1994usps}, SynthDigits \citep{ganin2015synthdigits-mnist-m}, MNIST-M \citep{ganin2015synthdigits-mnist-m}; each digit dataset represents a domain. Office-Caltech-10 consists of real-world object images from four domains: three domains (WebCam, DSLR, and Amazon) from Ofﬁce-31 dataset \citep{saenko2010office31} and one domain (Caltech) from Caltech-256 dataset \citep{griffin2007caltech}. DomainNet \citep{peng2019domainnet} contains large-sized 244x244 object images in six domains: Clipart, Infograph, Painting, Quickdraw, Real, and Sketch. To mimic the realistic scenarios where clients may not collect a large volume of data, we use a subset of standard digits datasets (7,438 training samples for each dataset instead of tens of thousands) as adopted in \cite{li2021fedbn}. We evenly split samples of each dataset into 
20 clients for cross-device FL with a total of 100 clients.
Similarly, we tailor the DomainNet dataset to include only 10 classes of 2,000-5,000 images. To simulate multi-domain FL, we construct a client to contain images 
from a single domain.

\paragraph{Implementation Details.} We implement FedWon using PyTorch \citep{paszke2017pytorch} and run experiments on a cluster of eight NVIDIA T4 GPUs. We evaluate the algorithms using three architectures: 6-layer convolution neural network (CNN) \citep{li2021fedbn} for Digits-Five dataset, AlexNet \citep{krizhevsky2017alexnet} and ResNet-18 \citep{he2016resnet} 
for Office-Caltech-10 dataset, and AlexNet \citep{krizhevsky2017alexnet} for DomainNet dataset. We use cross-entropy loss and stochastic gradient optimization (SGD) as the optimizer with learning rates tuned over the range of [0.001, 0.1] for all methods. Based on SGD, we adopt adaptive gradient clipping (AGC) that is specially designed for normalization-free networks \citep{brock2021high}. More details are provided in the supplementary.

\subsection{Performance Evaluation}
\label{sec:multi-domain-experiments}

We compare the performance of our proposed FedWon with the following three types of methods: (1) state-of-the-art methods that employ customized approaches on BN, including SiloBN \citep{andreux2020siloed}, FedBN \citep{li2021fedbn}, and FixBN \citep{zhong2023making}; (2) baseline algorithms, including FedProx \citep{fedprox}, FedAvg \citep{fedavg}, and SingleSet (i.e. training a model independently in each client with a single dataset); (3) alternative normalization methods, including FedAvg+GN and FedAvg+LN that replace BN layers with GN and LN layers, respectively.

Table \ref{tab:cross-silo-mean} presents a comprehensive comparison of the aforementioned methods under cross-silo FL on Digits-Five, Office-Caltech-10, and DomainNet datasets. Our proposed FedWon outperforms the state-of-the-art methods on most of the domains across all datasets. Specifically, FedProx, which adds a proximal term based on FedAvg, performs similarly to FedAvg. These two methods are better than SingleSet in Digits-Five dataset, but they may exhibit inferior performance compared to SingleSet in certain domains on the other two more challenging datasets. SiloBN and FixBN perform similarly to FedAvg, in terms of average accuracy; they are not primarily designed for multi-domain FL and are only capable of achieving the baseline results. In contrast, FedBN is specifically designed to excel in multi-domain FL and outperforms these methods.

\begin{wraptable}{r}{6.6cm}
   \vspace{-0.4cm}
   \caption{Performance comparison using small batch sizes \(B = \{1, 2\}\) on Office-Caltech-10 dataset. The abbreviations A, C, D, and W respectively represent 4 domains: Amazon, Caltech, DSLR, and WebCam. Our proposed FedWon achieves outstanding performance compared to existing methods.}\label{tab:small-batch-office}
   \begingroup
   \setlength{\tabcolsep}{0.47em}    
   \renewcommand{\arraystretch}{0.95}
     \begin{tabular}{l|l|ccccc}\toprule
     B &Methods &A &C &D &W \\\midrule
   \multirow{3}{*}{1} &FedAvg+GN &60.4 &52.0 &87.5 &84.8 \\
   &FedAvg+LN &55.7 &43.1 &84.4 &88.1 \\
   &\textbf{FedWon} &\textbf{66.7} &\textbf{55.1} &\textbf{96.9} &\textbf{89.8} \\\midrule
   \multirow{7}{*}{2} &FedAvg &64.1 &49.3 &87.5 &89.8 \\
   &FedAvg+GN &63.5 &52.0 &81.3 &84.8 \\
   &FedAvg+LN &58.3 &44.9 &87.5 &86.4 \\
   &FixBN &66.2 &50.7 &87.5 &88.1 \\
   &SiloBN &61.5 &47.1 &87.5 &86.4 \\
   &FedBN &59.4 &48.0 &96.9 &86.4 \\
   &\textbf{FedWon} &\textbf{66.2} &\textbf{54.7} &\textbf{93.8} &\textbf{89.8} \\
   \bottomrule
   \end{tabular}
   \endgroup
   \vspace{-0.5cm}
   \renewcommand{\arraystretch}{1}
 \end{wraptable} 

Besides, we discover that simply replacing BN with GN (FedAvg+GN) can boost the performance of FedAvg as GN does not depend on the batch statistics specific to domains; FedAvg+GN achieves comparable results as FedBN on Digits-Five and Office-Caltech-10 datasets. 
Notably, our proposed FedWon surpasses both FedAvg+GN and FedBN in terms of the average accuracy on all datasets. Although FedWon falls slightly behind FedBN by less than 1\% on two domains across these datasets, it outperforms FedBN by more than 17\% on certain domains. 
These results demonstrate the effectiveness of FedWon under the cross-silo FL scenario. We report the mean of three runs of experiments here and results with standard deviation in Table \ref{tab:cross-silo} in the Appendix.

 \begin{figure}[t!]
   \centering
   \begin{subfigure}[t]{0.5\textwidth}
    \includegraphics[width=\textwidth]{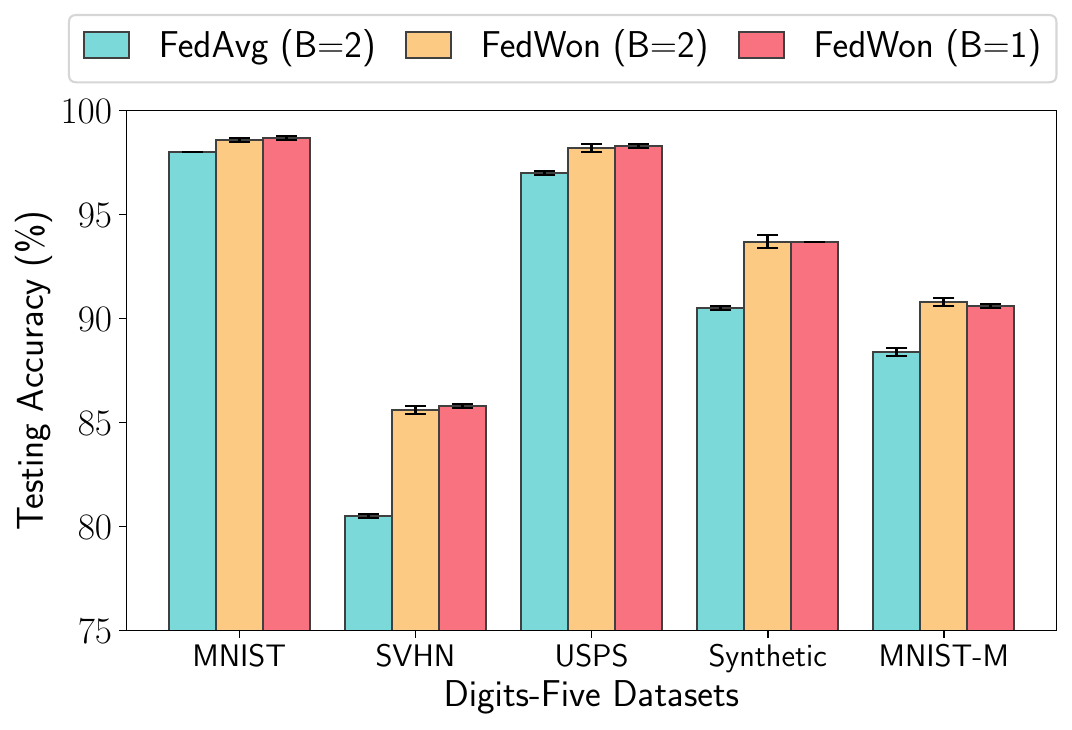}
    \label{fig:small_batch_digits}
  \end{subfigure}
  \hfill
  \begin{subfigure}[t]{0.45\textwidth}
       \includegraphics[width=\textwidth]{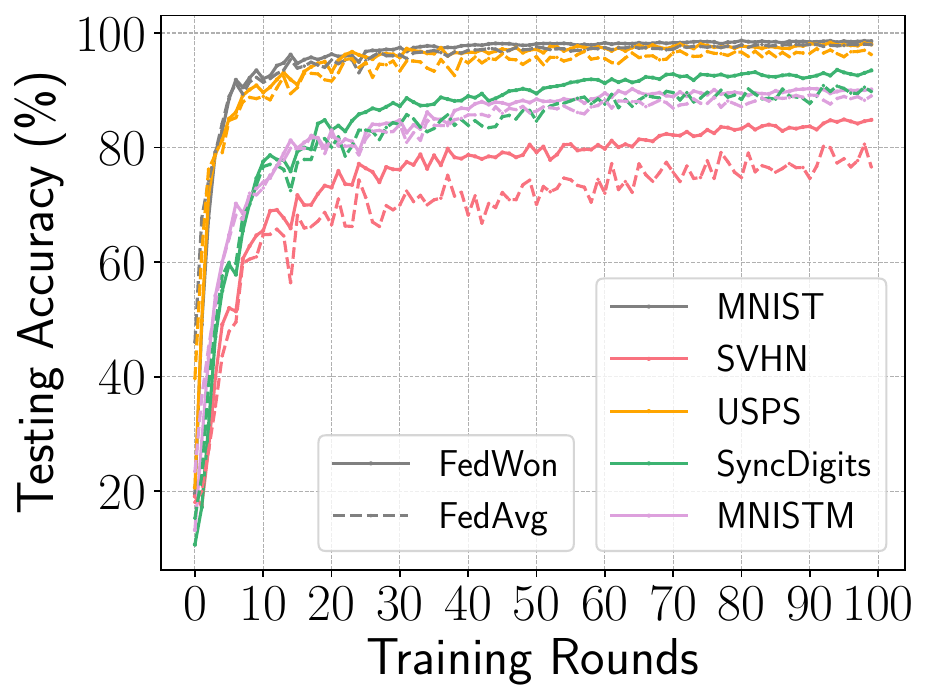}
       \label{fig:training-curve}
   \end{subfigure}
   \vspace{-0.5cm}
  \caption{Testing accuracy comparison of FedWon and FedAvg on Digits-Five dataset. Left: comparison of performance using small batch sizes B =\{1, 2\}, where 10 out of 100 clients are randomly selected to train in each round. Right: comparison of testing accuracy over the course of training with randomly selected 10 out of a total of 100 clients and batch size B = 2.}
  \label{fig:mnist}
 \end{figure}

\textbf{Effectiveness on Small Batch Size.} Table \ref{tab:small-batch-office} compares the performance of our proposed FedWon with state-of-the-art methods using small batch sizes \(B = \{1, 2\}\) on Office-Caltech-10 dataset.
FedWon achieves outstanding performance, with competitive results even at a batch size of 1. While FedAvg+GN and FedAvg+LN also achieve comparable results on batch size $B = 1$, they require additional computational cost during inference to calculate the running mean and variance, whereas our method does not have such constraints and achieves even better performance. The capability of our method to perform well under small batch sizes is particularly important for cross-device FL, as some edge devices may only be capable of training with small batch sizes under constrained resources.
We have fine-tuned the learning rates for all methods and reported the best ones.

\begin{wrapfigure}{r}{5.6cm}
  \vspace{-0.4cm}
  \centering
  \begin{subfigure}[t]{0.4\textwidth}
     \centering
      \includegraphics[width=\textwidth]{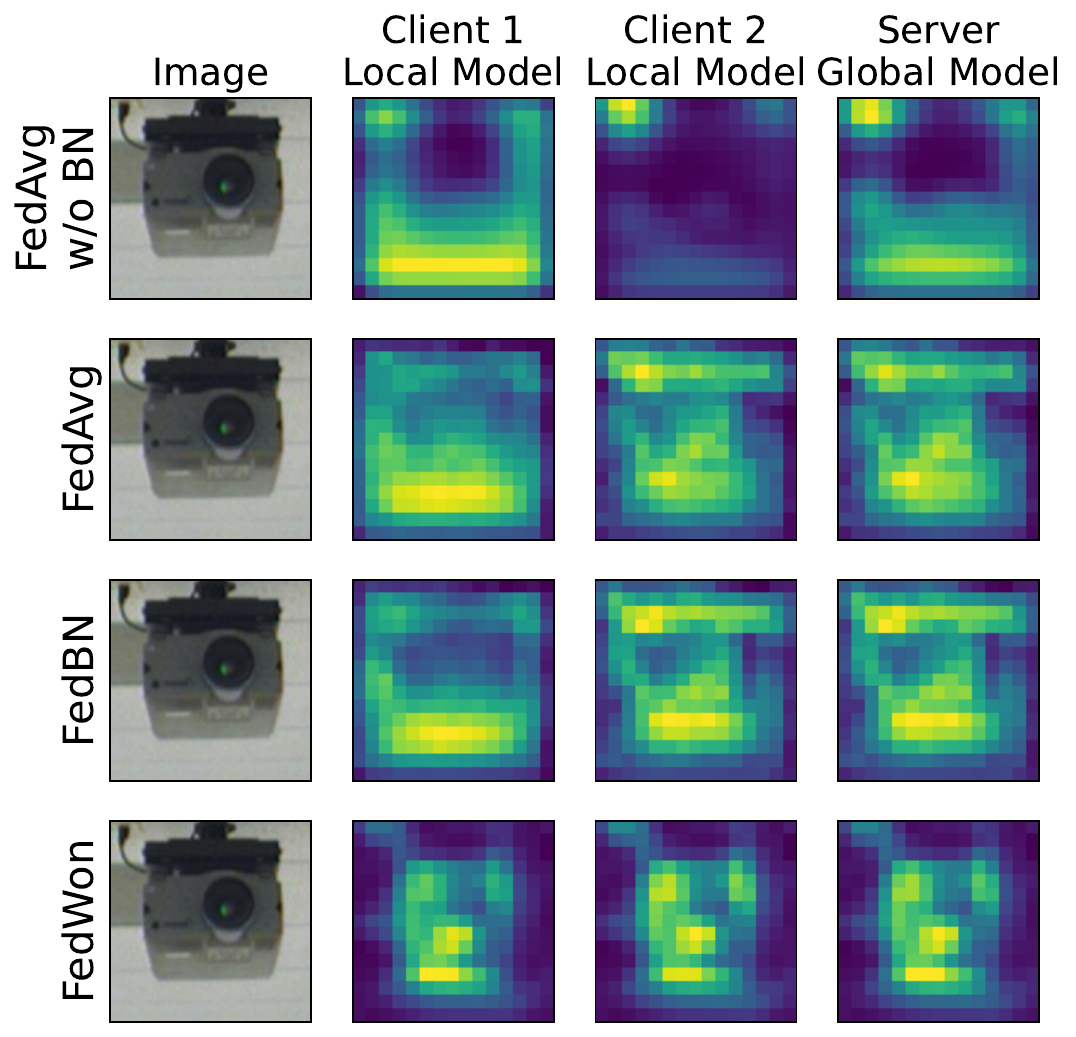}
      \label{fig:feature-map}
     \end{subfigure}
  \hfill
  \begin{subfigure}[t]{0.4\textwidth}
     \centering
     \begingroup
     \vspace{-0.2cm}
     \begin{tabular}{lcc}\toprule
        Methods &C$\leftrightarrow$C &S$\leftrightarrow$C \\\midrule
        FedAvg w/o BN &0.811 &0.873 \\
        FedAvg &0.865 &0.869 \\
        FedBN &0.901 &0.929 \\
        FedWon (Ours) &\textbf{0.993} &\textbf{0.997} \\
        \bottomrule
        \end{tabular}
     \endgroup
      \label{tab:cosine-similairty}
  \end{subfigure}     
 \caption{Analysis of feature maps with the Caltech-10 dataset. Top: visualization of feature maps of the last convolution layer. Bottom: comparison on average cosine similarity of feature maps between client (C$\leftrightarrow$C), and between a client and server (S$\leftrightarrow$C).}
 \label{fig:visualization}
 \vspace{-0.6cm}
\end{wrapfigure}

\textbf{Cross-device FL with Small Batch Size and Client Selection.} We assess the impact of randomly selecting a fraction of clients to participate in training in each round, which is common in cross-device FL where not all clients join in training.  We conduct experiments with fraction \(C = \{0.1, 0.2\} \) out of 100 clients on Digits-Five dataset, i.e., \(K = \{10, 20\} \) clients are selected to participate in training in each round. Table \ref{tab:client-fraction} shows that the performance of our FedWon is better than FedAvg under all client fractions. FedBN is not compared as it is not applicable in cross-device FL. We also evaluate small batch sizes in cross-device FL, with $K = 10$ clients selected per round. Figure \ref{fig:mnist} (left) shows that the performance of FedAvg degrades with batch size $B=2$, while our proposed FedWon with batch sizes \(B=\{1, 2\}\) achieves consistently comparable results to running with larger batch sizes. Besides, Figure \ref{fig:mnist} (right) shows the changes in testing accuracy over the course of training. It indicates that FedWon achieves better convergence speed without BN.

\begingroup
\setlength{\tabcolsep}{0.44em}
\begin{table}[t]\centering
  \vspace{-0.1cm}
  \caption{Testing accuracy comparison on randomly selecting a fraction \(C = \{0.1, 0.2\} \) out of a total of 100 clients for training each round with batch size $B = 4$ on Digits-Five dataset. FedWon consistently outperforms FedAvg. We report the mean (standard deviation) of three runs of experiments.}\label{tab:client-fraction}
  \begin{tabular}{c|l|cccccc}\toprule
  C &Method &MNIST &SVHN &USPS &SynthDigits &MNIST-M &\cellcolor[HTML]{efefef}Average \\\midrule
  \multirow{2}{*}{0.1} &FedAvg &98.2 {\footnotesize (0.4)} &81.0 {\footnotesize (0.7)} &97.2 {\footnotesize (0.5)} &91.6 {\footnotesize (1.6)} &89.3 {\footnotesize (0.5)} &\cellcolor[HTML]{efefef}91.5 {\footnotesize (0.8)} \\
  &\textbf{FedWon {\footnotesize (Ours)} }&\textbf{98.6 {\footnotesize (0.1)}} &\textbf{85.4 {\footnotesize (0.3)}} &\textbf{98.3 {\footnotesize (0.2)}} &\textbf{93.6 {\footnotesize (0.2)}} &\textbf{90.5 {\footnotesize (0.3)}} &\cellcolor[HTML]{efefef}\textbf{93.3 {\footnotesize (0.1)}} \\\midrule
  \multirow{2}{*}{0.2} &FedAvg &97.9 {\footnotesize (0.1)} &80.2 {\footnotesize (0.0)} &97.0 {\footnotesize (0.1)} &91.2 {\footnotesize (0.0)} &89.3 {\footnotesize (0.0)} &\cellcolor[HTML]{efefef}91.1 {\footnotesize (0.0)} \\
  &\textbf{FedWon {\footnotesize (Ours)} }&\textbf{98.7 {\footnotesize (0.1)}} &\textbf{86.0 {\footnotesize (0.3)}} &\textbf{98.2 {\footnotesize (0.2)}} &\textbf{94.1 {\footnotesize (0.2)}} &\textbf{90.8 {\footnotesize (0.1)}} &\cellcolor[HTML]{efefef}\textbf{93.6 {\footnotesize (0.1)}} \\
  \bottomrule
  \end{tabular}
\end{table}
\endgroup

\textbf{Visualization and Analysis of Feature Maps.} We aim to further study the reason behind the superior performance of FedWon. Figure \ref{fig:visualization} (top) visualizes feature maps of the last convolution layer of two client local models and one server global model on the Office-Caltech-10 dataset. The feature maps of FedAvg without (w/o) BN have a limited focus on the object of interest. While FedAvg and FedBN perform better, their feature maps display noticeable disparities between client local models.

In contrast, FedWon showcases superior feature map visualizations, with subtle differences observed among feature maps from different models. To provide further insight, we present the average cosine similarity of all feature maps between client local models (C$\leftrightarrow$C) and between the server global model and a client local model (S$\leftrightarrow$C) in Figure \ref{fig:visualization} (bottom).
These results demonstrate the effectiveness of FedWon, which achieves high similarity scores, approaching the maximum value of 1. This finding suggests that FedWon excels at effectively mitigating domain shifts across different domains. Building upon these insights, we extend our analysis to demonstrate that FedWon exhibits superior domain adaptation and generalization capabilities empirically in Table \ref{tab:domain-generalization} in the Appendix.

We also demonstrate that FedWon achieves significantly superior performance on medical diagnosis in Appendix \ref{sec:medical}, which is encouraging and shows the potential of FedWon in the healthcare field.

 \subsection{Ablation Studies}

 \begin{wraptable}{r}{7.1cm}  
     \vspace{-0.4cm}
     \caption{Ablation studies on the impact of WSConv on Caltech-10 dataset. It significantly boosts performance on both batch sizes B = 32 and B = 2.}\label{tab:ablations}
     \centering
     \begingroup
    \renewcommand{\arraystretch}{1}

     \begin{tabular}{c|c|cccc}\toprule
       B &WSConv &A &C &D &W \\\midrule
       \multirow{2}{*}{32} &$\checkmark$ &\textbf{63.7} &\textbf{51.0} &\textbf{96.3} &\textbf{91.2}\\
       & &46.4 &37.3 &68.8 &71.2  \\\midrule
       \multirow{2}{*}{2} &$\checkmark$ &\textbf{67.2} &\textbf{55.6} &\textbf{96.9} &\textbf{93.2} \\
       & &54.7 &44.0 &84.4 &78.0  \\
       \bottomrule
     \end{tabular}
     \endgroup
     \vspace{-0.4cm}
 \end{wraptable}

We conduct ablation studies to further analyze the impact of WSConv at batch sizes $B = 32$ and $B = 2$ on the Office-Caltech-10 dataset. 
 Table \ref{tab:ablations} compares the performance with and without WSConv after removing all normalization layers. It demonstrates that replacing convolution layers with WSConv significantly enhances performance. These experiments use a learning rate of $\eta = 0.08$ for $B = 32$ and $\eta = 0.01$ for $B = 2$. We provide more experiment details in the Appendix \ref{sec:exp-suppl}.

\section{Experiments on Skewed Label Distribution}

This section extends evaluation from multi-domain FL to skewed label distribution. We demonstrate that our proposed FedWon is also effective in addressing this problem.

\textbf{Dataset and Implementation.} We simulate skewed label distribution using CIFAR-10 dataset \citep{cifar10-2009}, which comprises 50,000 training samples and 10,000 testing samples. We split training samples into 100 clients and construct i.i.d data and three different levels of label skewness using Dirichlet process Dir($\alpha$) with \(\alpha = \{0.1, 0.5, 1\}\), where Dir(0.1) is the most heterogeneous setting. We run experiments using MobileNetV2 \citep{sandler2018mobilenetv2} with a fraction $C = 0.1$ randomly selected clients (i.e., $K = 10$)  out of a total of 100 clients in each round.

\begin{figure}[h]
  \centering
  \begin{subfigure}[t]{0.6\textwidth}
    \begin{tabular}{l|c|c|c|c}\toprule
      Methods &i.i.d &Dir (1) &Dir (0.5) &Dir (0.1) \\\midrule
      FedAvg &75.0 &64.5 &61.1 &36.0 \\
      FedAvg+GN &65.3 &58.8 &51.8 &21.5 \\
      FedAvg+LN &69.2 &61.8 &57.9 &23.3 \\
      FixBN &75.4 &64.1 &61.2 &34.7 \\
      \textbf{FedWon (Ours)} &\textbf{75.7} &\textbf{72.8} &\textbf{70.7} &\textbf{41.9} \\
      \bottomrule
      \end{tabular}
      \label{tab:label-shift}
  \end{subfigure}  
  \hfill
  \begin{subfigure}[t]{0.36\textwidth}
      \vspace{-1.5cm}
      \includegraphics[width=\textwidth]{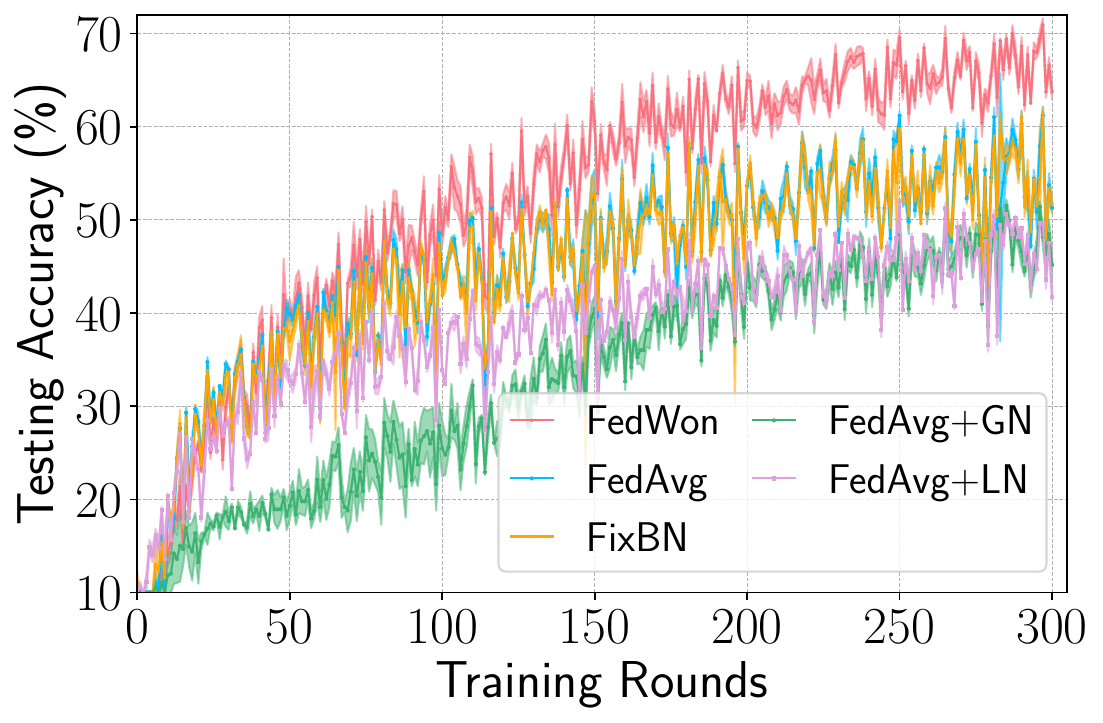}
      \label{fig:cifar-training}
  \end{subfigure}
  \vspace{-0.5cm}
 \caption{Testing accuracy comparison using MobileNetV2 as backbone on CIFAR-10 dataset. \textit{Left}: performance on different levels of label skewness, where Dir (0.1) represents the most skewed label distribution setting. \textit{Right}: changes in testing accuracy over the course of training on Dir (0.5).}
 \label{fig:skewed-label}
\end{figure}

\textbf{Performance Comparison.}
Figure \ref{fig:skewed-label} (left) compares FedWon with FedAvg, FedAvg+GN, FedAvg+LN, and FixBN. FedWon achieves similar performance as FedAvg and FixBN on the i.i.d setting, but outperforms all methods across different degrees of label skewness. We do not compare with FedBN and SiloBN as they are not suitable for cross-device FL and provide the comparison of cross-silo FL scenario in Table \ref{tab:cross-silo-skewed} in the Appendix.
Figure \ref{fig:skewed-label} (right) shows changes in testing accuracy over the course of training under the Dir (0.5) setting. FedWon converges to a better position than the other methods. These experiments indicate the possibility of employing our proposed FL without normalization to solve the skewed label distribution problem.

\section{Conclusion}

In conclusion, we propose FedWon, a new method for multi-domain FL by removing BN layers from DNNs and reparameterizing convolution layers with weight scaled convolution. Extensive experiments across four datasets and models demonstrate that this simple yet effective method outperforms state-of-the-art methods in a wide range of settings. Notably, FedWon is versatile for both cross-silo and cross-device FL. Its ability to train on small batch sizes is particularly useful for resource-constrained devices. 
Future work can conduct evaluations of this method under a broader range of datasets and backbones for skewed label distribution. Extending this paradigm from supervised to semi-supervised and unsupervised scenarios is also of interest.

\bibliography{iclr2024_conference}

\begin{thebibliography}{68}
\providecommand{\natexlab}[1]{#1}
\providecommand{\url}[1]{\texttt{#1}}
\expandafter\ifx\csname urlstyle\endcsname\relax
  \providecommand{\doi}[1]{doi: #1}\else
  \providecommand{\doi}{doi: \begingroup \urlstyle{rm}\Url}\fi

\bibitem[Andreux et~al.(2020)Andreux, du~Terrail, Beguier, and Tramel]{andreux2020siloed}
Mathieu Andreux, Jean~Ogier du~Terrail, Constance Beguier, and Eric~W. Tramel.
\newblock Siloed federated learning for multi-centric histopathology datasets.
\newblock In \emph{Domain Adaptation and Representation Transfer, and Distributed and Collaborative Learning}, pp.\  129--139. Springer International Publishing, 2020.
\newblock ISBN 978-3-030-60548-3.

\bibitem[Ba et~al.(2016)Ba, Kiros, and Hinton]{ba2016layer}
Jimmy~Lei Ba, Jamie~Ryan Kiros, and Geoffrey~E Hinton.
\newblock Layer normalization.
\newblock \emph{arXiv preprint arXiv:1607.06450}, 2016.

\bibitem[Bernecker et~al.(2022)Bernecker, Peters, Schlett, Bamberg, Theis, Rueckert, Wei{\ss}, and Albarqouni]{bernecker2022fednorm}
Tobias Bernecker, Annette Peters, Christopher~L Schlett, Fabian Bamberg, Fabian Theis, Daniel Rueckert, Jakob Wei{\ss}, and Shadi Albarqouni.
\newblock Fednorm: Modality-based normalization in federated learning for multi-modal liver segmentation.
\newblock \emph{arXiv preprint arXiv:2205.11096}, 2022.

\bibitem[Bjorck et~al.(2018)Bjorck, Gomes, Selman, and Weinberger]{bjorck2018understandingbn}
Nils Bjorck, Carla~P Gomes, Bart Selman, and Kilian~Q Weinberger.
\newblock Understanding batch normalization.
\newblock \emph{Advances in neural information processing systems}, 31, 2018.

\bibitem[Brock et~al.(2021{\natexlab{a}})Brock, De, and Smith]{brock2021characterizing}
Andrew Brock, Soham De, and Samuel~L Smith.
\newblock Characterizing signal propagation to close the performance gap in unnormalized resnets.
\newblock \emph{International Conference on Learning Representations}, 2021{\natexlab{a}}.

\bibitem[Brock et~al.(2021{\natexlab{b}})Brock, De, Smith, and Simonyan]{brock2021high}
Andy Brock, Soham De, Samuel~L Smith, and Karen Simonyan.
\newblock High-performance large-scale image recognition without normalization.
\newblock In \emph{International Conference on Machine Learning}, pp.\  1059--1071. PMLR, 2021{\natexlab{b}}.

\bibitem[Casella et~al.(2023)Casella, Esposito, Sciarappa, Cavazzoni, and Aldinucci]{casella2023experimenting}
Bruno Casella, Roberto Esposito, Antonio Sciarappa, Carlo Cavazzoni, and Marco Aldinucci.
\newblock Experimenting with normalization layers in federated learning on non-iid scenarios.
\newblock \emph{arXiv preprint arXiv:2303.10630}, 2023.

\bibitem[Chang et~al.(2019)Chang, You, Seo, Kwak, and Han]{chang2019domain}
Woong-Gi Chang, Tackgeun You, Seonguk Seo, Suha Kwak, and Bohyung Han.
\newblock Domain-specific batch normalization for unsupervised domain adaptation.
\newblock In \emph{Proceedings of the IEEE/CVF conference on Computer Vision and Pattern Recognition}, pp.\  7354--7362, 2019.

\bibitem[Chen et~al.(2022)Chen, Liu, Ma, and Lyu]{chen2022calfat}
Chen Chen, Yuchen Liu, Xingjun Ma, and Lingjuan Lyu.
\newblock Calfat: Calibrated federated adversarial training with label skewness.
\newblock \emph{Advances in Neural Information Processing Systems}, 2022.

\bibitem[Chen et~al.(2018)Chen, Zhou, Tang, Yang, Cao, and Gu]{chen2018closing}
Jinghui Chen, Dongruo Zhou, Yiqi Tang, Ziyan Yang, Yuan Cao, and Quanquan Gu.
\newblock Closing the generalization gap of adaptive gradient methods in training deep neural networks.
\newblock \emph{arXiv preprint arXiv:1806.06763}, 2018.

\bibitem[Codella et~al.(2018)Codella, Gutman, Celebi, Helba, Marchetti, Dusza, Kalloo, Liopyris, Mishra, Kittler, et~al.]{codella2018skin}
Noel~CF Codella, David Gutman, M~Emre Celebi, Brian Helba, Michael~A Marchetti, Stephen~W Dusza, Aadi Kalloo, Konstantinos Liopyris, Nabin Mishra, Harald Kittler, et~al.
\newblock Skin lesion analysis toward melanoma detection: A challenge at the 2017 international symposium on biomedical imaging (isbi), hosted by the international skin imaging collaboration (isic).
\newblock In \emph{2018 IEEE 15th international symposium on biomedical imaging (ISBI 2018)}, pp.\  168--172. IEEE, 2018.

\bibitem[Combalia et~al.(2019)Combalia, Codella, Rotemberg, Helba, Vilaplana, Reiter, Carrera, Barreiro, Halpern, Puig, et~al.]{combalia2019bcn20000}
Marc Combalia, Noel~CF Codella, Veronica Rotemberg, Brian Helba, Veronica Vilaplana, Ofer Reiter, Cristina Carrera, Alicia Barreiro, Allan~C Halpern, Susana Puig, et~al.
\newblock Bcn20000: Dermoscopic lesions in the wild.
\newblock \emph{arXiv preprint arXiv:1908.02288}, 2019.

\bibitem[Cordts et~al.(2016)Cordts, Omran, Ramos, Rehfeld, Enzweiler, Benenson, Franke, Roth, and Schiele]{cordts2016cityscapes}
Marius Cordts, Mohamed Omran, Sebastian Ramos, Timo Rehfeld, Markus Enzweiler, Rodrigo Benenson, Uwe Franke, Stefan Roth, and Bernt Schiele.
\newblock The cityscapes dataset for semantic urban scene understanding.
\newblock In \emph{Proceedings of the IEEE conference on computer vision and pattern recognition}, pp.\  3213--3223, 2016.

\bibitem[De \& Smith(2020)De and Smith]{de2020batch}
Soham De and Sam Smith.
\newblock Batch normalization biases residual blocks towards the identity function in deep networks.
\newblock \emph{Advances in Neural Information Processing Systems}, 33:\penalty0 19964--19975, 2020.

\bibitem[Du et~al.(2022)Du, Sun, Li, Chen, Zhang, Li, and Chen]{du2022rethinking}
Zhixu Du, Jingwei Sun, Ang Li, Pin-Yu Chen, Jianyi Zhang, Hai"~Helen" Li, and Yiran Chen.
\newblock Rethinking normalization methods in federated learning.
\newblock In \emph{Proceedings of the 3rd International Workshop on Distributed Machine Learning}, pp.\  16--22, 2022.

\bibitem[Ganin \& Lempitsky(2015)Ganin and Lempitsky]{ganin2015synthdigits-mnist-m}
Yaroslav Ganin and Victor Lempitsky.
\newblock Unsupervised domain adaptation by backpropagation.
\newblock In \emph{International conference on machine learning}, pp.\  1180--1189. PMLR, 2015.

\bibitem[Gong et~al.(2012)Gong, Shi, Sha, and Grauman]{gong2012office-caltech10}
Boqing Gong, Yuan Shi, Fei Sha, and Kristen Grauman.
\newblock Geodesic flow kernel for unsupervised domain adaptation.
\newblock In \emph{2012 IEEE conference on computer vision and pattern recognition}, pp.\  2066--2073. IEEE, 2012.

\bibitem[Griffin et~al.(2007)Griffin, Holub, and Perona]{griffin2007caltech}
Gregory Griffin, Alex Holub, and Pietro Perona.
\newblock Caltech-256 object category dataset.
\newblock 2007.

\bibitem[Hanin \& Rolnick(2018)Hanin and Rolnick]{hanin2018start}
Boris Hanin and David Rolnick.
\newblock How to start training: The effect of initialization and architecture.
\newblock \emph{Advances in Neural Information Processing Systems}, 31, 2018.

\bibitem[Hard et~al.(2018)Hard, Rao, Mathews, Ramaswamy, Beaufays, Augenstein, Eichner, Kiddon, and Ramage]{hard2018gboard}
Andrew Hard, Kanishka Rao, Rajiv Mathews, Swaroop Ramaswamy, Fran{\c{c}}oise Beaufays, Sean Augenstein, Hubert Eichner, Chlo{\'e} Kiddon, and Daniel Ramage.
\newblock Federated learning for mobile keyboard prediction.
\newblock \emph{arXiv preprint arXiv:1811.03604}, 2018.

\bibitem[He et~al.(2016)He, Zhang, Ren, and Sun]{he2016resnet}
Kaiming He, Xiangyu Zhang, Shaoqing Ren, and Jian Sun.
\newblock Deep residual learning for image recognition.
\newblock In \emph{Proceedings of the IEEE conference on computer vision and pattern recognition}, pp.\  770--778, 2016.

\bibitem[Hsieh et~al.(2020)Hsieh, Phanishayee, Mutlu, and Gibbons]{hsieh2020non}
Kevin Hsieh, Amar Phanishayee, Onur Mutlu, and Phillip Gibbons.
\newblock The non-iid data quagmire of decentralized machine learning.
\newblock In \emph{International Conference on Machine Learning}, pp.\  4387--4398. PMLR, 2020.

\bibitem[Hull(1994)]{hull1994usps}
Jonathan~J. Hull.
\newblock A database for handwritten text recognition research.
\newblock \emph{IEEE Transactions on pattern analysis and machine intelligence}, 16\penalty0 (5):\penalty0 550--554, 1994.

\bibitem[Ioffe(2017)]{ioffe2017batch}
Sergey Ioffe.
\newblock Batch renormalization: Towards reducing minibatch dependence in batch-normalized models.
\newblock \emph{Advances in neural information processing systems}, 30, 2017.

\bibitem[Ioffe \& Szegedy(2015)Ioffe and Szegedy]{ioffe2015bn}
Sergey Ioffe and Christian Szegedy.
\newblock Batch normalization: Accelerating deep network training by reducing internal covariate shift.
\newblock In \emph{International conference on machine learning}, pp.\  448--456. pmlr, 2015.

\bibitem[Kairouz et~al.(2021)Kairouz, McMahan, Avent, Bellet, Bennis, Bhagoji, Bonawitz, Charles, Cormode, Cummings, et~al.]{kairouz2021advances}
Peter Kairouz, H~Brendan McMahan, Brendan Avent, Aur{\'e}lien Bellet, Mehdi Bennis, Arjun~Nitin Bhagoji, Kallista Bonawitz, Zachary Charles, Graham Cormode, Rachel Cummings, et~al.
\newblock Advances and open problems in federated learning.
\newblock \emph{Foundations and Trends{\textregistered} in Machine Learning}, 14\penalty0 (1--2):\penalty0 1--210, 2021.

\bibitem[Karimireddy et~al.(2020)Karimireddy, Kale, Mohri, Reddi, Stich, and Suresh]{karimireddy2020scaffold}
Sai~Praneeth Karimireddy, Satyen Kale, Mehryar Mohri, Sashank Reddi, Sebastian Stich, and Ananda~Theertha Suresh.
\newblock Scaffold: Stochastic controlled averaging for federated learning.
\newblock In \emph{International Conference on Machine Learning}, pp.\  5132--5143. PMLR, 2020.

\bibitem[Krizhevsky et~al.(2009)Krizhevsky, Hinton, et~al.]{cifar10-2009}
Alex Krizhevsky, Geoffrey Hinton, et~al.
\newblock Learning multiple layers of features from tiny images.
\newblock 2009.

\bibitem[Krizhevsky et~al.(2017)Krizhevsky, Sutskever, and Hinton]{krizhevsky2017alexnet}
Alex Krizhevsky, Ilya Sutskever, and Geoffrey~E Hinton.
\newblock Imagenet classification with deep convolutional neural networks.
\newblock \emph{Communications of the ACM}, 60\penalty0 (6):\penalty0 84--90, 2017.

\bibitem[LeCun et~al.(1998)LeCun, Bottou, Bengio, and Haffner]{lecun1998mnist}
Yann LeCun, L{\'e}on Bottou, Yoshua Bengio, and Patrick Haffner.
\newblock Gradient-based learning applied to document recognition.
\newblock \emph{Proceedings of the IEEE}, 86\penalty0 (11):\penalty0 2278--2324, 1998.

\bibitem[Li et~al.(2020{\natexlab{a}})Li, Sahu, Talwalkar, and Smith]{Li2020FedChallenges}
Tian Li, Anit~Kumar Sahu, Ameet Talwalkar, and Virginia Smith.
\newblock Federated learning: Challenges, methods, and future directions.
\newblock \emph{IEEE Signal Processing Magazine}, 37:\penalty0 50--60, 2020{\natexlab{a}}.

\bibitem[Li et~al.(2020{\natexlab{b}})Li, Sahu, Zaheer, Sanjabi, Talwalkar, and Smith]{fedprox}
Tian Li, Anit~Kumar Sahu, Manzil Zaheer, Maziar Sanjabi, Ameet Talwalkar, and Virginia Smith.
\newblock Federated optimization in heterogeneous networks.
\newblock \emph{Proceedings of Machine Learning and Systems}, 2:\penalty0 429--450, 2020{\natexlab{b}}.

\bibitem[Li et~al.(2019)Li, Milletar{\`\i}, Xu, Rieke, Hancox, Zhu, Baust, Cheng, Ourselin, Cardoso, et~al.]{li2019brain-tumor1}
Wenqi Li, Fausto Milletar{\`\i}, Daguang Xu, Nicola Rieke, Jonny Hancox, Wentao Zhu, Maximilian Baust, Yan Cheng, S{\'e}bastien Ourselin, M~Jorge Cardoso, et~al.
\newblock Privacy-preserving federated brain tumour segmentation.
\newblock In \emph{International Workshop on Machine Learning in Medical Imaging}, pp.\  133--141. Springer, 2019.

\bibitem[Li et~al.(2021)Li, Jiang, Zhang, Kamp, and Dou]{li2021fedbn}
Xiaoxiao Li, Meirui Jiang, Xiaofei Zhang, Michael Kamp, and Qi~Dou.
\newblock Fedbn: Federated learning on non-iid features via local batch normalization.
\newblock \emph{arXiv preprint arXiv:2102.07623}, 2021.

\bibitem[Li et~al.(2016)Li, Wang, Shi, Liu, and Hou]{li2016revisiting}
Yanghao Li, Naiyan Wang, Jianping Shi, Jiaying Liu, and Xiaodi Hou.
\newblock Revisiting batch normalization for practical domain adaptation.
\newblock \emph{arXiv preprint arXiv:1603.04779}, 2016.

\bibitem[Lin et~al.(2017)Lin, Goyal, Girshick, He, and Doll{\'a}r]{lin2017focal}
Tsung-Yi Lin, Priya Goyal, Ross Girshick, Kaiming He, and Piotr Doll{\'a}r.
\newblock Focal loss for dense object detection.
\newblock In \emph{Proceedings of the IEEE international conference on computer vision}, pp.\  2980--2988, 2017.

\bibitem[Lu et~al.(2022)Lu, Wang, Chen, Qin, Xu, Dimitriadis, and Qin]{lu2022personalized}
Wang Lu, Jindong Wang, Yiqiang Chen, Xin Qin, Renjun Xu, Dimitrios Dimitriadis, and Tao Qin.
\newblock Personalized federated learning with adaptive batchnorm for healthcare.
\newblock \emph{IEEE Transactions on Big Data}, 2022.

\bibitem[McMahan et~al.(2017)McMahan, Moore, Ramage, Hampson, and y~Arcas]{fedavg}
Brendan McMahan, Eider Moore, Daniel Ramage, Seth Hampson, and Blaise~Aguera y~Arcas.
\newblock Communication-efficient learning of deep networks from decentralized data.
\newblock In \emph{Artificial intelligence and statistics}, pp.\  1273--1282. PMLR, 2017.

\bibitem[Netzer et~al.(2011)Netzer, Wang, Coates, Bissacco, Wu, and Ng]{netzer2011svhn}
Yuval Netzer, Tao Wang, Adam Coates, Alessandro Bissacco, Bo~Wu, and Andrew~Y Ng.
\newblock Reading digits in natural images with unsupervised feature learning.
\newblock 2011.

\bibitem[Nguyen et~al.(2021)Nguyen, Do, Tran, Nguyen, Duong, Phan, Tjiputra, and Tran]{nguyen2021auto1}
Anh Nguyen, Tuong Do, Minh Tran, Binh~X Nguyen, Chien Duong, Tu~Phan, Erman Tjiputra, and Quang~D Tran.
\newblock Deep federated learning for autonomous driving.
\newblock \emph{arXiv preprint arXiv:2110.05754}, 2021.

\bibitem[Paszke et~al.(2017)Paszke, Gross, Chintala, Chanan, Yang, DeVito, Lin, Desmaison, Antiga, and Lerer]{paszke2017pytorch}
Adam Paszke, Sam Gross, Soumith Chintala, Gregory Chanan, Edward Yang, Zachary DeVito, Zeming Lin, Alban Desmaison, Luca Antiga, and Adam Lerer.
\newblock Automatic differentiation in pytorch.
\newblock 2017.

\bibitem[Paulik et~al.(2021)Paulik, Seigel, Mason, Telaar, Kluivers, van Dalen, Lau, Carlson, Granqvist, Vandevelde, et~al.]{paulik2021apple}
Matthias Paulik, Matt Seigel, Henry Mason, Dominic Telaar, Joris Kluivers, Rogier van Dalen, Chi~Wai Lau, Luke Carlson, Filip Granqvist, Chris Vandevelde, et~al.
\newblock Federated evaluation and tuning for on-device personalization: System design \& applications.
\newblock \emph{arXiv preprint arXiv:2102.08503}, 2021.

\bibitem[Peng et~al.(2019)Peng, Bai, Xia, Huang, Saenko, and Wang]{peng2019domainnet}
Xingchao Peng, Qinxun Bai, Xide Xia, Zijun Huang, Kate Saenko, and Bo~Wang.
\newblock Moment matching for multi-source domain adaptation.
\newblock In \emph{Proceedings of the IEEE/CVF international conference on computer vision}, pp.\  1406--1415, 2019.

\bibitem[Posner et~al.(2021)Posner, Tseng, Aloqaily, and Jararweh]{posner2021vehicular}
Jason Posner, Lewis Tseng, Moayad Aloqaily, and Yaser Jararweh.
\newblock Federated learning in vehicular networks: opportunities and solutions.
\newblock \emph{IEEE Network}, 35\penalty0 (2):\penalty0 152--159, 2021.

\bibitem[Saenko et~al.(2010)Saenko, Kulis, Fritz, and Darrell]{saenko2010office31}
Kate Saenko, Brian Kulis, Mario Fritz, and Trevor Darrell.
\newblock Adapting visual category models to new domains.
\newblock In \emph{Computer Vision--ECCV 2010: 11th European Conference on Computer Vision, Heraklion, Crete, Greece, September 5-11, 2010, Proceedings, Part IV 11}, pp.\  213--226. Springer, 2010.

\bibitem[Sandler et~al.(2018)Sandler, Howard, Zhu, Zhmoginov, and Chen]{sandler2018mobilenetv2}
Mark Sandler, Andrew Howard, Menglong Zhu, Andrey Zhmoginov, and Liang-Chieh Chen.
\newblock Mobilenetv2: Inverted residuals and linear bottlenecks.
\newblock In \emph{Proceedings of the IEEE conference on computer vision and pattern recognition}, pp.\  4510--4520, 2018.

\bibitem[Santurkar et~al.(2018)Santurkar, Tsipras, Ilyas, and Madry]{santurkar2018doesbn}
Shibani Santurkar, Dimitris Tsipras, Andrew Ilyas, and Aleksander Madry.
\newblock How does batch normalization help optimization?
\newblock \emph{Advances in neural information processing systems}, 31, 2018.

\bibitem[Shen et~al.(2022)Shen, Zhou, and Yu]{shen2022cd2}
Yiqing Shen, Yuyin Zhou, and Lequan Yu.
\newblock Cd2-pfed: Cyclic distillation-guided channel decoupling for model personalization in federated learning.
\newblock In \emph{Proceedings of the IEEE/CVF Conference on Computer Vision and Pattern Recognition}, pp.\  10041--10050, 2022.

\bibitem[Sun et~al.(2021)Sun, Huo, Yang, and Bai]{sun2021partialfed}
Benyuan Sun, Hongxing Huo, Yi~Yang, and Bo~Bai.
\newblock Partialfed: Cross-domain personalized federated learning via partial initialization.
\newblock \emph{Advances in Neural Information Processing Systems}, 34:\penalty0 23309--23320, 2021.

\bibitem[Tan \& Le(2019)Tan and Le]{tan2019efficientnet}
Mingxing Tan and Quoc Le.
\newblock Efficientnet: Rethinking model scaling for convolutional neural networks.
\newblock In \emph{International conference on machine learning}, pp.\  6105--6114. PMLR, 2019.

\bibitem[Tan et~al.(2023)Tan, Chen, Zhuang, Dong, Lyu, and Long]{tan2023heterogeneity}
Yue Tan, Chen Chen, Weiming Zhuang, Xin Dong, Lingjuan Lyu, and Guodong Long.
\newblock Is heterogeneity notorious? taming heterogeneity to handle test-time shift in federated learning.
\newblock In \emph{Thirty-seventh Conference on Neural Information Processing Systems}, 2023.

\bibitem[Terrail et~al.(2022)Terrail, Ayed, Cyffers, Grimberg, He, Loeb, Mangold, Marchand, Marfoq, Mushtaq, et~al.]{terrail2022flamby}
Jean Ogier~du Terrail, Samy-Safwan Ayed, Edwige Cyffers, Felix Grimberg, Chaoyang He, Regis Loeb, Paul Mangold, Tanguy Marchand, Othmane Marfoq, Erum Mushtaq, et~al.
\newblock Flamby: Datasets and benchmarks for cross-silo federated learning in realistic healthcare settings.
\newblock \emph{arXiv preprint arXiv:2210.04620}, 2022.

\bibitem[Tschandl et~al.(2018)Tschandl, Rosendahl, and Kittler]{tschandl2018ham10000}
Philipp Tschandl, Cliff Rosendahl, and Harald Kittler.
\newblock The ham10000 dataset, a large collection of multi-source dermatoscopic images of common pigmented skin lesions.
\newblock \emph{Scientific data}, 5\penalty0 (1):\penalty0 1--9, 2018.

\bibitem[Wang et~al.(2020)Wang, Yurochkin, Sun, Papailiopoulos, and Khazaeni]{wang2020fedma}
Hongyi Wang, Mikhail Yurochkin, Yuekai Sun, Dimitris Papailiopoulos, and Yasaman Khazaeni.
\newblock Federated learning with matched averaging.
\newblock In \emph{International Conference on Learning Representations}, 2020.
\newblock URL \url{https://openreview.net/forum?id=BkluqlSFDS}.

\bibitem[Wang et~al.(2023)Wang, Shi, and Chang]{wang2023batch}
Yanmeng Wang, Qingjiang Shi, and Tsung-Hui Chang.
\newblock Why batch normalization damage federated learning on non-iid data?
\newblock \emph{arXiv preprint arXiv:2301.02982}, 2023.

\bibitem[Wu \& He(2018)Wu and He]{wu2018group}
Yuxin Wu and Kaiming He.
\newblock Group normalization.
\newblock In \emph{Proceedings of the European conference on computer vision (ECCV)}, pp.\  3--19, 2018.

\bibitem[Yao et~al.(2022)Yao, Gong, Qi, Cui, Zhu, and Yang]{yao2022federated}
Chun-Han Yao, Boqing Gong, Hang Qi, Yin Cui, Yukun Zhu, and Ming-Hsuan Yang.
\newblock Federated multi-target domain adaptation.
\newblock In \emph{Proceedings of the IEEE/CVF Winter Conference on Applications of Computer Vision}, pp.\  1424--1433, 2022.

\bibitem[Yu et~al.(2020)Yu, Chen, Wang, Xian, Chen, Liu, Madhavan, and Darrell]{yu2020bdd100k}
Fisher Yu, Haofeng Chen, Xin Wang, Wenqi Xian, Yingying Chen, Fangchen Liu, Vashisht Madhavan, and Trevor Darrell.
\newblock Bdd100k: A diverse driving dataset for heterogeneous multitask learning.
\newblock In \emph{Proceedings of the IEEE/CVF conference on computer vision and pattern recognition}, pp.\  2636--2645, 2020.

\bibitem[Zhang et~al.(2019)Zhang, Dauphin, and Ma]{zhang2019fixup}
Hongyi Zhang, Yann~N Dauphin, and Tengyu Ma.
\newblock Fixup initialization: Residual learning without normalization.
\newblock \emph{arXiv preprint arXiv:1901.09321}, 2019.

\bibitem[Zhang et~al.(2021)Zhang, Bosch, and Olsson]{zhang2021auto}
Hongyi Zhang, Jan Bosch, and Helena~Holmstr{\"o}m Olsson.
\newblock End-to-end federated learning for autonomous driving vehicles.
\newblock In \emph{2021 International Joint Conference on Neural Networks (IJCNN)}, pp.\  1--8. IEEE, 2021.

\bibitem[Zhang et~al.(2023)Zhang, Chen, Zhuang, and Lv]{zhang2023addressing}
Jie Zhang, Chen Chen, Weiming Zhuang, and Lingjuan Lv.
\newblock Addressing catastrophic forgetting in federated class-continual learning.
\newblock \emph{arXiv preprint arXiv:2303.06937}, 2023.

\bibitem[Zhao et~al.(2018)Zhao, Li, Lai, Suda, Civin, and Chandra]{zhao2018non-iid}
Yue Zhao, Meng Li, Liangzhen Lai, Naveen Suda, Damon Civin, and Vikas Chandra.
\newblock Federated learning with non-iid data.
\newblock \emph{CoRR}, abs/1806.00582, 2018.
\newblock URL \url{http://arxiv.org/abs/1806.00582}.

\bibitem[Zhong et~al.(2023)Zhong, Chen, and Chao]{zhong2023making}
Jike Zhong, Hong-You Chen, and Wei-Lun Chao.
\newblock Making batch normalization great in federated deep learning.
\newblock \emph{arXiv preprint arXiv:2303.06530}, 2023.

\bibitem[Zhuang et~al.(2020)Zhuang, Wen, Zhang, Gan, Yin, Zhou, Zhang, and Yi]{zhuang2020fedreid}
Weiming Zhuang, Yonggang Wen, Xuesen Zhang, Xin Gan, Daiying Yin, Dongzhan Zhou, Shuai Zhang, and Shuai Yi.
\newblock Performance optimization of federated person re-identification via benchmark analysis.
\newblock In \emph{Proceedings of the 28th ACM International Conference on Multimedia}, pp.\  955--963, 2020.

\bibitem[Zhuang et~al.(2021)Zhuang, Gan, Wen, Zhang, and Yi]{zhuang2021fedu}
Weiming Zhuang, Xin Gan, Yonggang Wen, Shuai Zhang, and Shuai Yi.
\newblock Collaborative unsupervised visual representation learning from decentralized data.
\newblock In \emph{Proceedings of the IEEE/CVF International Conference on Computer Vision}, pp.\  4912--4921, 2021.

\bibitem[Zhuang et~al.(2022{\natexlab{a}})Zhuang, Gan, Wen, and Zhang]{zhuang2022easyfl}
Weiming Zhuang, Xin Gan, Yonggang Wen, and Shuai Zhang.
\newblock Easyfl: A low-code federated learning platform for dummies.
\newblock \emph{IEEE Internet of Things Journal}, 9\penalty0 (15):\penalty0 13740--13754, 2022{\natexlab{a}}.

\bibitem[Zhuang et~al.(2022{\natexlab{b}})Zhuang, Gan, Wen, and Zhang]{zhuang2022fedreid}
Weiming Zhuang, Xin Gan, Yonggang Wen, and Shuai Zhang.
\newblock Optimizing performance of federated person re-identification: Benchmarking and analysis.
\newblock \emph{ACM Transactions on Multimedia Computing, Communications, and Applications (TOMM)}, 2022{\natexlab{b}}.

\bibitem[Zhuang et~al.(2022{\natexlab{c}})Zhuang, Wen, and Zhang]{zhuang2022divergence}
Weiming Zhuang, Yonggang Wen, and Shuai Zhang.
\newblock Divergence-aware federated self-supervised learning.
\newblock \emph{International Conference on Learning Representations}, 2022{\natexlab{c}}.

\end{thebibliography}
\bibliographystyle{iclr2024_conference}

\newpage

\appendix
\section{Experimental Setup}

In this section, we provide more details of experimental setups, including datasets, model architectures, and implementation details.

\subsection{Dataets}

\begin{figure}[t]
  \centering
  \begin{subfigure}[t]{0.15\textwidth}
      \includegraphics[width=\textwidth]{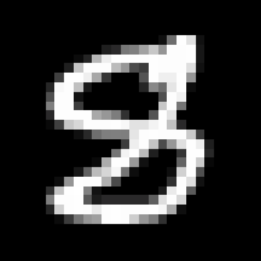}
      \caption{MNIST}
      \label{fig:mnist-data}
  \end{subfigure}
  \hfill
  \begin{subfigure}[t]{0.15\textwidth}
    \includegraphics[width=\textwidth]{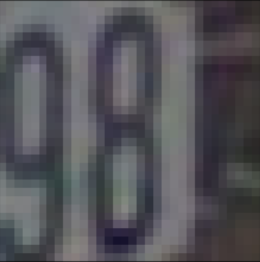}
    \caption{SVHN}
    \label{fig:svhn}
  \end{subfigure}
  \hfill
  \begin{subfigure}[t]{0.15\textwidth}
    \includegraphics[width=\textwidth]{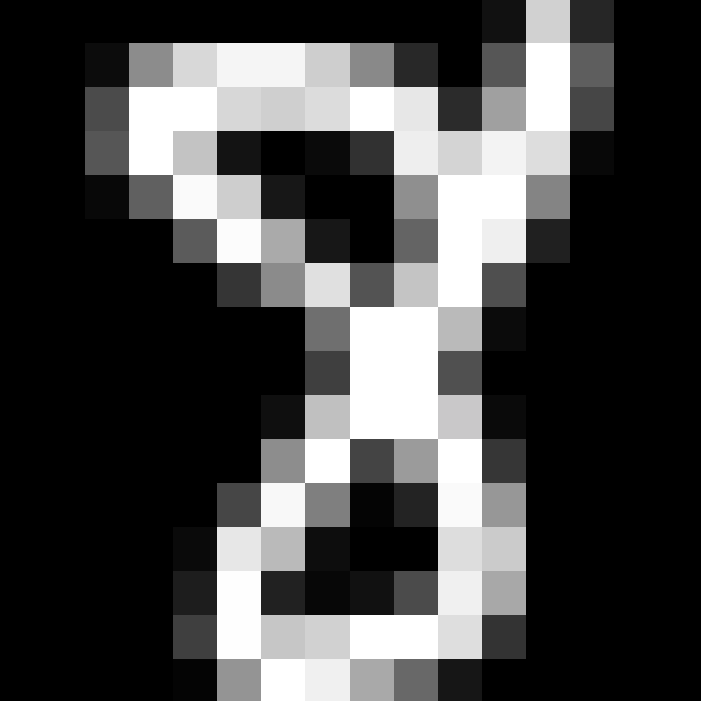}
    \caption{USPS}
    \label{fig:usps}
  \end{subfigure}
  \hfill
  \begin{subfigure}[t]{0.15\textwidth}
    \includegraphics[width=\textwidth]{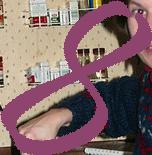}
    \caption{SynthDigits}
    \label{fig:synthdigits}
  \end{subfigure}
  \hfill
  \begin{subfigure}[t]{0.15\textwidth}
    \includegraphics[width=\textwidth]{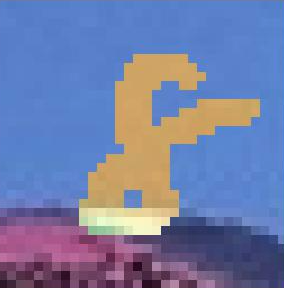}
    \caption{MNIST-M}
    \label{fig:mnist-m}
  \end{subfigure}  
 \caption{Visualization of samples from Digits-Five dataset.}
 \label{fig:digits-five}
\end{figure}

\begin{figure}[t]
  \centering
  \begin{subfigure}[t]{0.15\textwidth}
      \includegraphics[width=\textwidth]{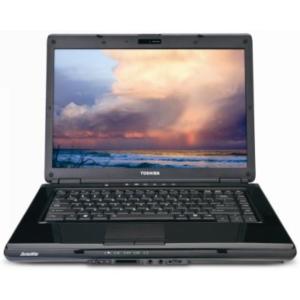}
      \caption{Amazon}
      \label{fig:amazon}
  \end{subfigure}
  \hfill
  \begin{subfigure}[t]{0.15\textwidth}
    \includegraphics[width=\textwidth]{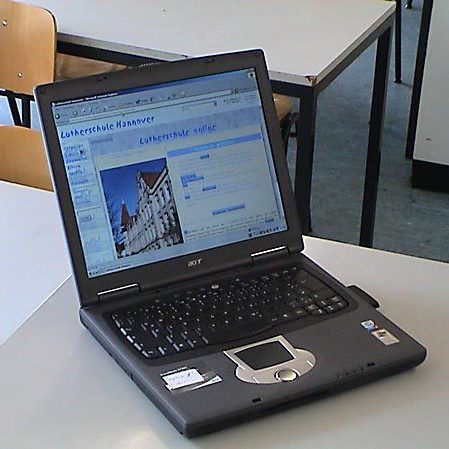}
    \caption{Caltech}
    \label{fig:caltech}
  \end{subfigure}
  \hfill
  \begin{subfigure}[t]{0.15\textwidth}
    \includegraphics[width=\textwidth]{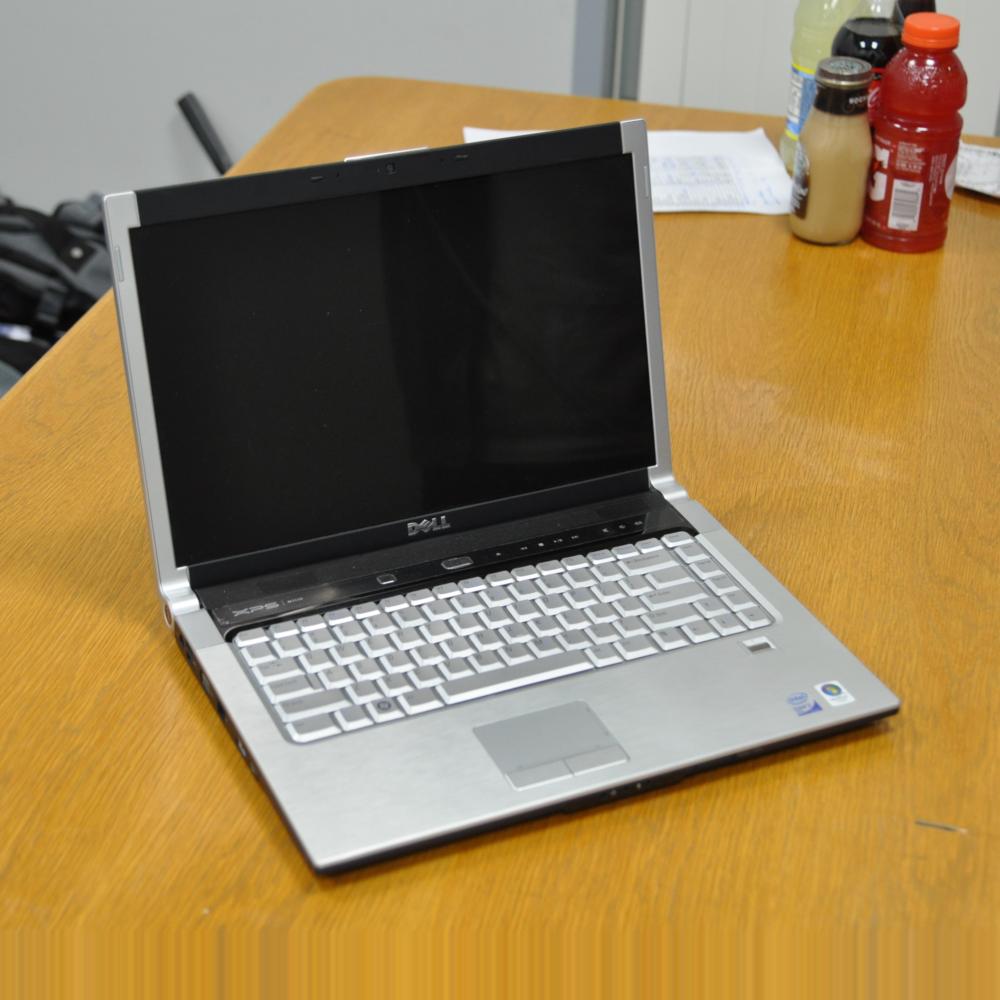}
    \caption{DSLR}
    \label{fig:dslr}
  \end{subfigure}
  \hfill
  \begin{subfigure}[t]{0.15\textwidth}
    \includegraphics[width=\textwidth]{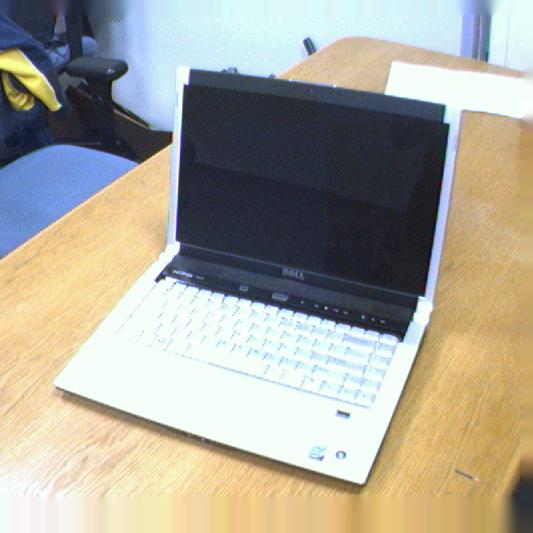}
    \caption{Webcam}
    \label{fig:webcam}
  \end{subfigure}
 \caption{Visualization of samples from Office-Caltech-10 dataset.}
 \label{fig:office-caltech}
\end{figure}

\begin{figure}[t!]
  \centering
  \begin{subfigure}[t]{0.15\textwidth}
      \includegraphics[width=\textwidth]{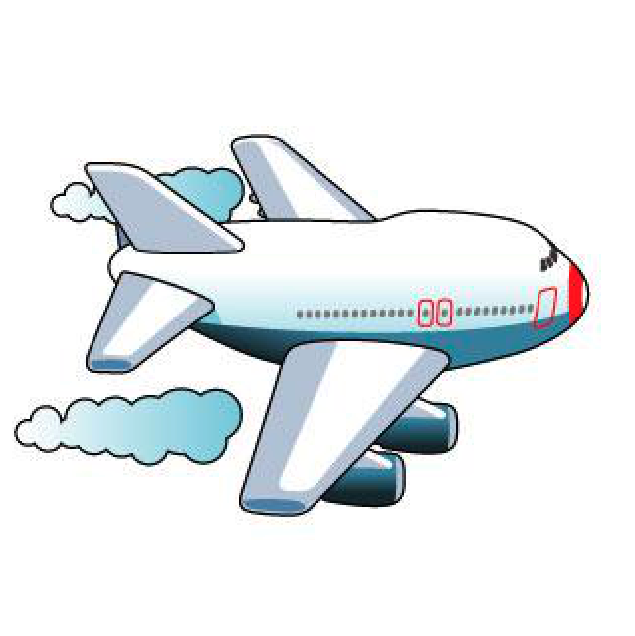}
      \caption{Clipart}
      \label{fig:clipart}
  \end{subfigure}
  \hfill
  \begin{subfigure}[t]{0.15\textwidth}
    \includegraphics[width=\textwidth]{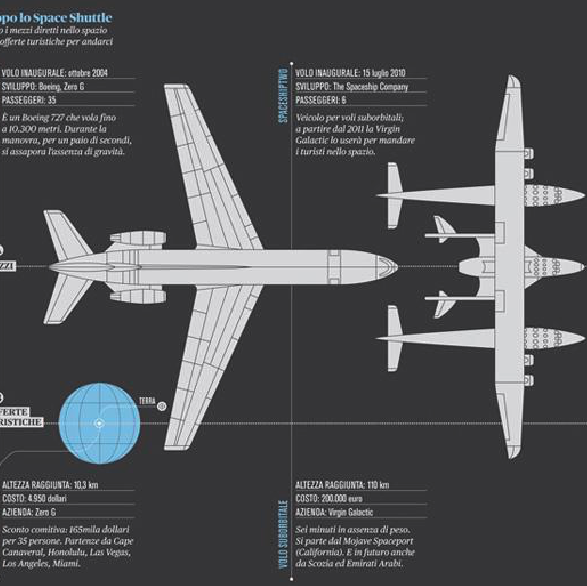}
    \caption{Infograph}
    \label{fig:infograph}
  \end{subfigure}
  \hfill
  \begin{subfigure}[t]{0.15\textwidth}
    \includegraphics[width=\textwidth]{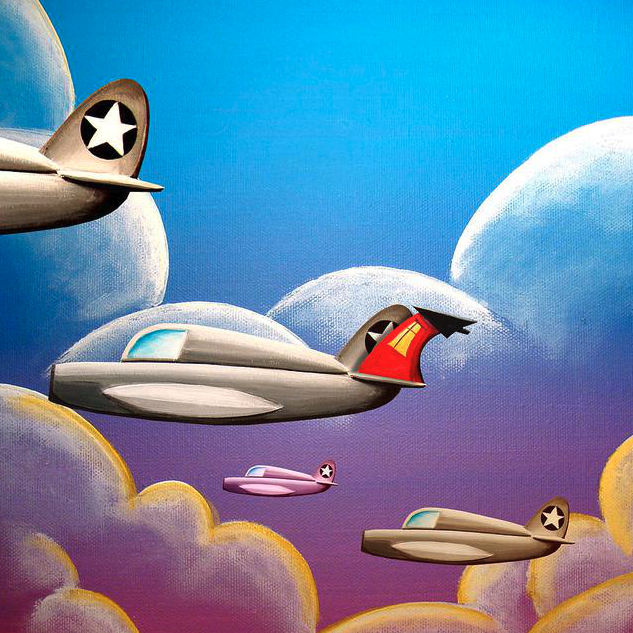}
    \caption{Painting}
    \label{fig:painting}
  \end{subfigure}
  \hfill
  \begin{subfigure}[t]{0.15\textwidth}
    \includegraphics[width=\textwidth]{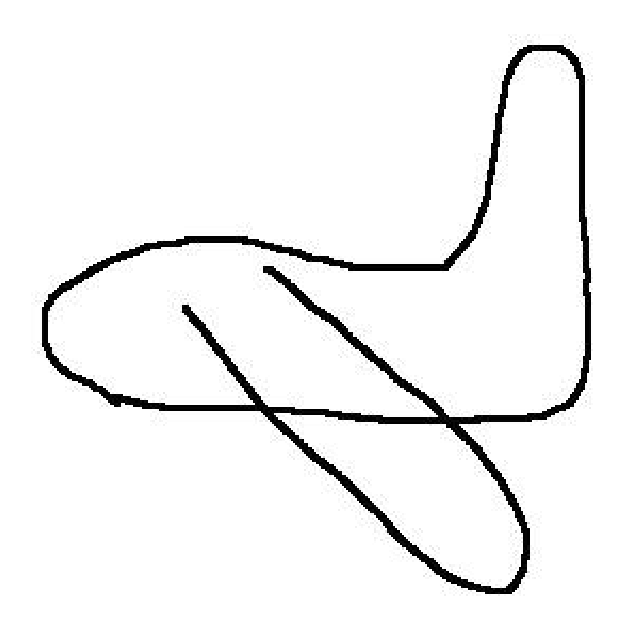}
    \caption{Quickdraw}
    \label{fig:quickdraw}
  \end{subfigure}
  \hfill
  \begin{subfigure}[t]{0.15\textwidth}
    \includegraphics[width=\textwidth]{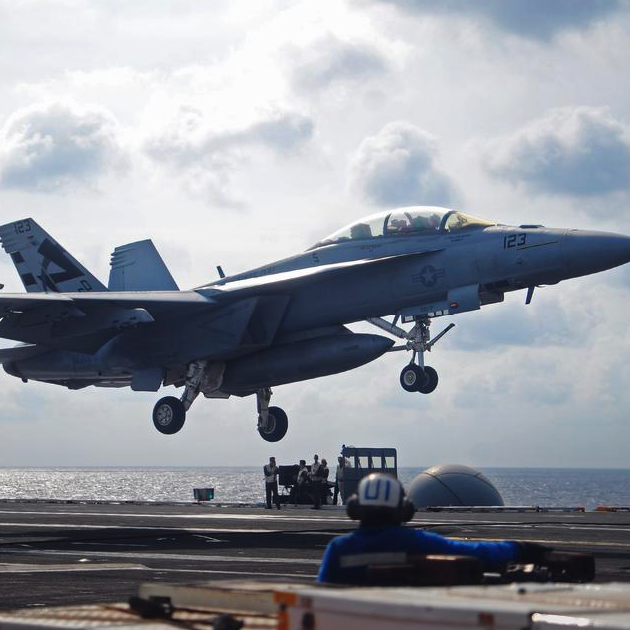}
    \caption{Real}
    \label{fig:real}
  \end{subfigure}  
  \hfill
  \begin{subfigure}[t]{0.15\textwidth}
    \includegraphics[width=\textwidth]{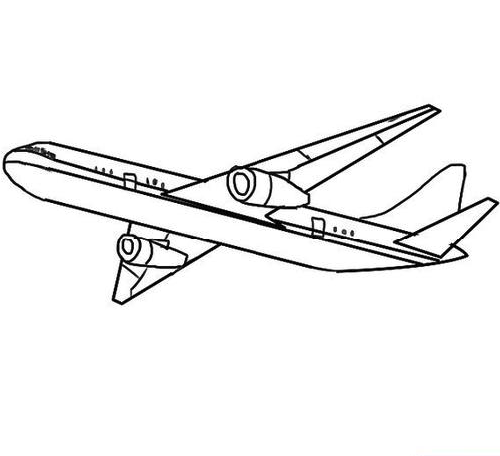}
    \caption{Sketch}
    \label{fig:sketch}
  \end{subfigure}    
 \caption{Visualization of samples from DomainNet dataset.}
 \label{fig:domainnet}
\end{figure}

Figure \ref{fig:digits-five}, \ref{fig:office-caltech}, and \ref{fig:domainnet} visualize three multi-domain datasets used in this work; these three datasets are Digits-Five \citep{li2021fedbn}, Office-Caltech-10 \citep{gong2012office-caltech10}, and DomainNet \citep{peng2019domainnet}, respectively. It shows that images under each dataset have significant domain gaps. We construct multi-domain FL by constraining each FL client to contain samples of the same domain. Each image is a sample from one client. Each FL client contains images of a dataset (domain). We follow FedBN \citep{li2021fedbn} to preprocess and transform these datasets.

\subsection{Model Architectures} 
\label{sec:arch}

Table \ref{tab:cnn} illustrates the model architectures for experiments on the Digits-Five dataset and Table \ref{tab:alexnet} illustrates the model architectures for experiments on Office-Caltech-10 and DomainNet datasets. For the convolution layer (Conv2D), the hyperparameters are in the sequence of input dimension, output dimension, kernel size, stride, and padding. For the max pooling layer (MaxPool2D), the hyperparameters are kernel and stride. For the fully connected layer (FC), the hyperparameters are input and output dimensions. For the batch normalization (BN) layer, the hyperparameter is the number of channels. For group normalization, the hyperparameters are the number of groups and the number of channels. FedAvg+LN shares a similar model architecture as FedAvg+GN but sets the number of groups to 1. The methods with BN are Standalone, FedAvg \citep{fedavg}, FedProx \citep{fedprox}, SiloBN \citep{andreux2020siloed}, FedBN \citep{li2021fedbn}, and FixBN \citep{zhong2023making}. These methods share the same model architecture. Note that the model architecture is not exactly the same as the ones used in FedBN \citep{li2021fedbn}, where they use a one-dimension BN layer as regularizer between FC layers but we use Dropout such that the comparisons are fair in terms of model architectures. 

Besides, we use the default implementation of ResNet-18 \citep{he2016resnet} and MobileNetV2 \citep{paszke2017pytorch} in PyTorch \citep{paszke2017pytorch} for methods with BN on the Office-Caltech-10 dataset and CIFAR-10 dataset, respectively. FedWon replaces the convolution layers in ResNet-18 and MobileNetV2 with WSConv and removes all batch normalization layers. FedAvg+GN and FedAvg+LN replace BN layers with GN layers. Specifically, FedAvg+GN sets the number of groups to 32 by default, but sets it to 8 when the output dimension is smaller than 32, and to 24 when the output dimension is 144 (to ensure divisibility); FedAvg+LN sets the number of groups in GN to 1. The source code will be released.

\subsection{Implementation and Training Details}

Listing \ref{list:wsconv} provides the implementation of WSConv in PyTorch. We employ the architectures described in Section \ref{sec:arch} to implement FedWon, adhering to the client training and server aggregation protocols of FedAvg \citep{fedavg}. We implement FedWon based on both EasyFL \citep{zhuang2022easyfl} for skewed label distribution experiments and FedBN original implementation for multi-domain FL experiments. For the implementation of FedBN, we reference the open-source code available in Github \footnote{https://github.com/med-air/FedBN}. To implement SiloBN \citep{andreux2020siloed}, we modify the FedBN implementation to aggregate only the BN parameters while keeping the BN statistics local. Unfortunately, as the source code for FixBN \citep{zhong2023making} is not publicly available, we implement it based on the description provided in the paper.

Besides, we summarize the compared algorithms in Table \ref{tab:compared-methods}.

\newpage

\begin{lstlisting}[language=Python, caption={WSConv implementation in PyTorch.}, label = {list:wsconv}]
  class WSConv(nn.Conv2d):
    def __init__(self,in_channels,out_channels,kernel_size,stride=1,padding=0,dilation=1,groups=1,bias=True,padding_mode='zeros'):
      super(WSConv, self).__init__(in_channels,out_channels,kernel_size,stride,padding,dilation,groups,bias,padding_mode)
      nn.init.xavier_normal_(self.weight)
      self.gain = nn.Parameter(torch.ones(self.out_channels, 1, 1, 1))
      _eps = torch.tensor(1e-4, requires_grad=False)
      _fan_in = torch.tensor(self.weight.shape[1:].numel(), requires_grad=False).type_as(self.weight)
      self.register_buffer('eps', _eps, persistent=False)
      self.register_buffer('fan_in', _fan_in, persistent=False)
  
    def standardized_weights(self):
      mean = torch.mean(self.weight, axis=[1,2,3], keepdims=True)
      var = torch.var(self.weight, axis=[1,2,3], keepdims=True)
      scale = torch.rsqrt(torch.maximum(var * self.fan_in, self.eps))
      return (self.weight - mean) * scale * self.gain
        
    def forward(self, x):
      return F.conv2d(
        input=x,
        weight=self.standardized_weights(),
        bias=self.bias,
        stride=self.stride,
        padding=self.padding,
        dilation=self.dilation,
        groups=self.groups
      )
  \end{lstlisting}

\begin{table}[t!]\centering
  \caption{Model architectures of Six-layer CNN for experiments on Digits-Five dataset.}\label{tab:cnn}
  \begin{tabular}{cllll}\toprule
    Layer &Methods with BN &FedWon &FedAvg+GN \\\midrule
    \multirow{3}{*}{1} &Conv2D(3, 64, 5, 1, 2) &\textbf{WSConv2D}(3, 64, 5, 1, 2) &Conv2D(3, 64, 5, 1, 2) \\
    &\textbf{BN}(64), ReLU &ReLU &\textbf{GN}(32, 64), ReLU \\
    &MaxPool2D(2, 2) &MaxPool2D(2, 2) &MaxPool2D(2, 2) \\\midrule
    \multirow{3}{*}{2} &Conv2D(64, 64, 5, 1, 2) &\textbf{WSConv2D}(64, 64, 5, 1, 2) &Conv2D(64, 64, 5, 1, 2) \\
    &\textbf{BN}(64), ReLU &ReLU &\textbf{GN}(32, 64), ReLU \\
    &MaxPool2D(2, 2) &MaxPool2D(2, 2) &MaxPool2D(2, 2) \\\midrule
    \multirow{2}{*}{3} &Conv2D(64, 128, 5, 1, 2) &\textbf{WSConv2D}(64, 128, 5, 1, 2) &Conv2D(64, 128, 5, 1, 2) \\
    &\textbf{BN}(128), ReLU &ReLU &\textbf{GN}(64, 128), ReLU \\\midrule
    \multirow{2}{*}{4} &Dropout, FC(6272, 2048) &Dropout, FC(6272, 2048) &Dropout, FC(6272, 2048) \\
    &ReLU &ReLU &ReLU \\\midrule
    \multirow{2}{*}{5} &Dropout, FC(2048, 512) &Dropout, FC(2048, 512) &Dropout, FC(2048, 512) \\
    &ReLU &ReLU &ReLU \\\midrule
    6 &FC(512, 10) &FC(512, 10) &FC(512, 10) \\
    \bottomrule
    \end{tabular}
\end{table}  

\begingroup
\setlength{\tabcolsep}{0.5em}
\begin{table}[h]\centering
  \caption{Model architectures of AlexNet for experiments on Office-Caltech-10 and DomainNet datasets.}\label{tab:alexnet}
  \begin{tabular}{cllll}\toprule
  Layer &Methods with BN &FedWon &FedAvg+GN \\\midrule
  \multirow{3}{*}{1} &Conv2D(3, 64, 11, 4, 2) &\textbf{WSConv2D}(3, 64, 11, 4, 2) &Conv2D(3, 64, 11, 4, 2) \\
  &\textbf{BN}(64), ReLU &ReLU &\textbf{GN}(32, 64), ReLU \\
  &MaxPool2D(3, 2) &MaxPool2D(3, 2) &MaxPool2D(3, 2) \\\midrule
  \multirow{3}{*}{2} &Conv2D(64, 192, 5, 1, 2) &\textbf{WSConv2D}(64, 192, 5, 1, 2) &Conv2D(64, 192, 5, 1, 2) \\
  &\textbf{BN}(192), ReLU &ReLU &\textbf{GN}(32, 192), ReLU \\
  &MaxPool2D(3, 2) &MaxPool2D(3, 2) &MaxPool2D(3, 2) \\\midrule
  \multirow{2}{*}{3} &Conv2D(192, 384, 3, 1, 1) &Conv2D(192, 384, 3, 1, 1) &Conv2D(192, 384, 3, 1, 1) \\
  &\textbf{BN}(384), ReLU &ReLU &\textbf{GN}(64,384), ReLU \\\midrule
  \multirow{2}{*}{4} &Conv2D(384, 256, 3, 1, 1) &\textbf{WSConv2D}(384, 256, 3, 1, 1) &Conv2D(384, 256, 3, 1, 1) \\
  &\textbf{BN}(256), ReLU &ReLU &\textbf{GN}(64, 256), ReLU \\\midrule
  \multirow{3}{*}{5} &Conv2D(256, 256, 3, 1, 1) &\textbf{WSConv2D}(256, 256, 3, 1, 1) &Conv2D(256, 256, 3, 1, 1) \\
  &\textbf{BN}(256), ReLU &ReLU &\textbf{GN}(64, 256), ReLU \\
  &MaxPool2D(3, 2) &MaxPool2D(3, 2) &MaxPool2D(3, 2) \\\midrule
  6 &AdaptiveAvgPool2D(6, 6) &AdaptiveAvgPool2D(6, 6) &AdaptiveAvgPool2D(6, 6) \\\midrule
  \multirow{2}{*}{7} &Dropout, FC(9216, 4096) &Dropout, FC(9216, 4096) &Dropout, FC(9216, 4096) \\
  &ReLU &ReLU &ReLU \\\midrule
  \multirow{2}{*}{8} &Dropout, FC(4096, 4096) &Dropout, FC(4096, 4096) &Dropout, FC(4096, 4096) \\
  &ReLU &ReLU &ReLU \\\midrule
  9 &FC(4096, 10) &FC(4096, 10) &FC(4096, 10) \\
  \bottomrule
  \end{tabular}
\end{table}
\endgroup

 \begingroup
 \setlength{\tabcolsep}{0.3em}
\begin{table}[t]\centering
  \caption{Summary of compared methods on different aspects. \checkmark and $\times$ means the method supports and does not support the attribute, respectively. $\bigcirc$ means that no prior studies are conducted to analyze whether the method supports the attribute.}
  \label{tab:compared-methods}
  \resizebox{\textwidth}{!}{
   \begin{tabular}{lcccccc}\toprule
  Method &Has no BN &Multi-domain FL &Skewed Labeld Distribution &Cross-silo FL &Cross-device FL \\\midrule
  FedAvg & $\times$ & \checkmark & \checkmark & \checkmark & \checkmark \\
  FedAvg+GN & \checkmark & $\bigcirc$ & \checkmark & \checkmark & \checkmark \\
  FedAVG+LN & \checkmark & $\bigcirc$ & \checkmark & \checkmark & \checkmark \\
  FixBN & $\times$ & $\times$ & \checkmark & \checkmark & \checkmark \\
  SiloBN & $\times$ & $\bigcirc$ & \checkmark & \checkmark & $\times$ \\
  FedBN & $\times$ & \checkmark & $\bigcirc$ & \checkmark & $\times$ \\
  FedWon & \checkmark & \checkmark & \checkmark & \checkmark & \checkmark \\
  \bottomrule
  \end{tabular}
  }
 \end{table}
\endgroup

By default, we conduct experiments with local epochs $E = 1$ and batch size $B = 32$ across all datasets. Stochastic gradient optimization (SGD) is used as the optimizer, with learning rates tuned in the range of [0.001, 0.1] for all methods.  Specifically, for FedWon experiments with a batch size of $B = 32$, we incorporate adaptive gradient clipping (AGC) \citep{brock2021high}, which is specifically designed for normalization-free networks. AGC applies gradient clipping to the weight matrix $W^l \in \mathbb{R}^{N \times M}$ of the $l^{th}$ layer, where the gradient $G^l \in \mathbb{R}^{N \times M}$ is clipped with a threshold $\lambda$ before updating the model. The clipping operation for each row $i$ of $G^l$ can be expressed as follows:
\begin{equation}
  G_i^l=
  \begin{cases}
    \lambda \frac{||W_i^l||_F^*}{||G_i^l||_F} G_i^l, & \text{if}\ \frac{||G_i^l||_F}{||W_i^l||_F^*} > \lambda, \\
    G_i^l, & \text{otherwise},
  \end{cases}
\end{equation}
where $||\cdot||_F$ is the the Frobenius norm, i.e. $||W^l||_F = \sqrt{\sum_i^N \sum_j^M(W_{i,j})^2)}$, $||W_i^l||_F^* =$ max$(||W_i||_F, \epsilon)$ with default $\epsilon = 1e-3$. We only use AGC for FedWon with batch size $B = 32$ and bypass AGC on small batch sizes such as $B = \{1, 2, 4\}$. The impact of AGC and the clipping threshold is further analyzed in Section \ref{sec:exp-suppl}. 

We tune the learning rates for the methods compared in the main manuscript and provide their specific learning rates below. Table \ref{tab:main-lr} illustrates the learning rates of different methods on three datasets, corresponding to the experiments of Table 1 in the main manuscript. We use a clipping threshold of 0.64 for Digits-Five, 1.28 for Office-Caltech-10, and 1.28 for the DomainNet dataset. Additionally, Table \ref{tab:small-bs-lr} presents the learning rates used for experiments with small batch sizes $B = \{1, 2, 4\}$ on Office-Caltech-10 and Digits-Five datasets. Table \ref{tab:resnet18-lr} displays the learning rates used for experiments using ResNet-18 as the backbone. All experiments on Digits-Five are trained for 100 rounds and experiments on Office-Caltech-10 and DomainNet are trained for 300 rounds. 

For evaluation of skewed label distribution, all experiments are run with local epoch $E = 5$ for 300 rounds. We use SGD as the optimizer and tune the learning in the range of [0.001, 0.1] for different algorithms. 

\begingroup
\setlength{\tabcolsep}{0.4em}
\begin{table}[!htp]\centering
  \caption{Learning rates of different methods in the experiments of Table 1 in the manuscript.}\label{tab:main-lr}
  \begin{tabular}{l|ccccccccc}\toprule
  Datasets &Standalone &FedAvg &FedProx &$^a$+GN &$^b$+LN &SiloBN &FixBN &FedBN &Ours \\\midrule
  Digits-Five &0.1 &0.1 &0.1 &0.1 &0.1 &0.1 &0.1 &0.1 &0.05 \\
  Caltech-10 &0.01 &0.01 &0.01 &0.01 &0.01 &0.01 &0.01 &0.01 &0.1 \\
  DomainNet &0.01 &0.01 &0.01 &0.01 &0.01 &0.01 &0.01 &0.05 &0.05 \\
  \bottomrule
  \end{tabular}
  \footnotesize{$^a$+GN means FedAvg+GN, $^b$+LN means FedAvg+LN}
\end{table}
\endgroup

\begin{table}[!htp]\centering
  \caption{Learning rates of experiments on small batch sizes. Left: learning rates of experiments of small batch sizes $B = \{1, 2, 4\}$ on Office-Caltech-10 dataset. Right: learning rates of experiments of small batch sizes of randonly selecting a fraction $C = \{0.1, 0.2, 0.4\}$ out of total clients on Digits-Five dataset.}\label{tab:small-bs-lr}
  \begin{subtable}{0.7\textwidth}
    \begingroup
    \setlength{\tabcolsep}{0.23em}
    \begin{tabular}{l|cccccccc}\toprule
      B &FedAvg &SiloBN &FixBN &FedBN &FedAvg+GN &FedAvg+LN &Ours \\\midrule
      1 &- &- &- &- &0.001 &0.001 &0.005 \\
      2 &0.001 &0.001 &0.001 &0.001 &0.001 &0.001 &0.01 \\
      4 &0.001 &0.01 &0.01 &0.01 &0.001 &0.001 &0.03 \\
      \bottomrule
      \end{tabular}    
    \endgroup
  \end{subtable}
  \hfill
  \begin{subtable}{0.25\textwidth}
    \begingroup
    \setlength{\tabcolsep}{0.3em}
    \begin{tabular}{lcccc}\toprule
      C &B &FedAvg &Ours \\\midrule
      0.1 &1 &- &0.01 \\
      0.1 &2 &0.005 &0.01 \\
      0.1 &4 &0.01 &0.04 \\
      0.2 &4 &0.01 &0.04 \\
      0.4 &4 &0.01 &0.04 \\
      \bottomrule
    \end{tabular}
    \endgroup
  \end{subtable}
\end{table}

\begingroup
\begin{table}[!htp]\centering
  \caption{Learning rates $\eta$ of different methods in the experiments of using ResNet-20 as backbone.}\label{tab:resnet18-lr}
  \begin{tabular}{cccccccc}\toprule
   &FedAvg &FedAvg+GN &FedAvg+LN &SiloBN &FixBN &FedBN &Ours \\\midrule
  $\eta$ &0.1 &0.03 &0.01 &0.05 &0.03 &0.03 &0.1 \\
  \bottomrule
  \end{tabular}
\end{table}
\endgroup

\section{Experiments}
\label{sec:exp-suppl}

This section provides more experiment results that provide further insights into the behavior of FedWon and shed light on the effects of different parameters.

\subsection{Experiments on Medical Images}
\label{sec:medical}

To further study how our proposed FedWon benefits multi-domain FL in real-world scenarios, we extend evaluation to diagnosis of skin lesions using datasets from ISIC2019 Challenge \citep{codella2018skin,combalia2019bcn20000} and the HAM10000 \citep{tschandl2018ham10000} dataset. The dataset contains images collected from four hospitals, where one hospital with 3 different imaging technologies. We follow Flamby \citep{terrail2022flamby} to construct them as six different centers: BCN, Vidir-molemax, Vidir-modern, Rosendahl, MSK, and Vienna-dias, with each center's images representing a unique domain. In total, the dataset encompasses 23,247 images of skin lesions, including 9930 training samples and 2483 testing samples from BCN; 3163 training samples and 791 testing samples from Vidir-molemax, 2691 training samples and 672 testing samples from Vidir-modern, 1807 training samples and 452 testing samples from Rosendahl, 655 training samples and 164 testing samples from MSK, and 351 training samples and 88 samples from Vienna-dias. In the experimental setup, we simulate the scenarios where multiple healthcare centers collaborate to train a skin lesion diagnosis model, with each client representing a healthcare center. The task is to conduct image classification for 8 different melanoma classes.

We run the experiments using ResNet-18 \citep{he2016resnet} (without any pre-training) with local epoch $E = 1$ and batch size $B = 64$ for 50 rounds. We use SGD optimizer with learning rate $\eta = 0.005$ for FedAvg and FedWon and $\eta = 0.001$ for FedBN. The learning rate is tuned among $\{0.001, 0.005, 0.01, 0.05\}$. We follow the implementation in Flamby \footnote{https://github.com/owkin/FLamby/} to use a weighted focal loss \citep{lin2017focal} and data augmentations.

Table \ref{tab:isic} shows the testing accuracy of FedAvg, FedBN, and our proposed FedWon across the six healthcare center domains. In this challenging setting, FedBN only achieves similar performance to FedAvg. In contrast, FedWon outperforms both FedAvg and FedBN in all domains by a significant margin. The results are inspiring and demonstrates the potential of deploying FedWon to healthcare application scenarios, where data is often scarce, isolated, and spans multiple domains.

\begin{table}[t]\centering
  \caption{Evaluation on Fed-ISIC2019 dataset with medical images from six different centers. FedWon outperforms FedAvg and FedBN by a signifcant margin in all domains.}\label{tab:isic}
  \begin{tabular}{lccccccc}\toprule
  Methods &Center 1 &Center 2 &Center 3 &Center 4 &Center 5 &Center 6 \\\midrule
  FedAvg &0.40 &0.21 &0.37 &0.42 &0.39 &0.43 \\
  FedBN &0.31 &0.38 &0.43 &0.39 &0.30 &0.36 \\
  \textbf{FedWon (Ours)} &\textbf{0.46} &\textbf{0.43} &\textbf{0.52} &\textbf{0.56} &\textbf{0.40} &\textbf{0.59} \\
  \bottomrule
  \end{tabular}
\end{table}

\begin{table}[t!]\centering
  \caption{Comparison of methods on domain generalization capability using the Office-Caltech-10 dataset, where we employ Amazon, Caltech, and DSLR as the seen domains during training and WebCam as the unseen domain for evaluation.}
   \label{tab:domain-generalization}
   \begin{tabular}{lcccccc}\toprule
   \multirow{2}{*}{Methods} &\multicolumn{3}{c}{Seen Domains} &&\multicolumn{1}{c}{Unseen Domain} \\
   \cmidrule(lr){2-4} \cmidrule(lr){6-6}
   &Amazon &Caltech &DSLR && WebCam \\\midrule
   FedAvg w/o BN &38.0 &33.7 &40.6 &&28.8 \\
   FedAvg &58.9 &42.7 &59.4 &&52.5 \\
   FedBN &65.6 &48.0 &78.1 &&61.0 \\
   \textbf{FedWon (Ours)} &\textbf{66.1} &\textbf{51.6} &\textbf{90.6} &&\textbf{67.8} \\
   \bottomrule
   \end{tabular}
\end{table}

  \begin{table}[t]
     \centering
     \caption{Testing accuracy (\%) comparison using ResNet-20 on Office-Caltech-10 Dataset.}\label{tab:office-resnet}
     \begingroup
     \setlength{\tabcolsep}{0.43em}
     \begin{tabular}{l|cccccc}\toprule
     Methods &Amazon &Caltech &DSLR &WebCam &\cellcolor[HTML]{efefef}Avg \\\midrule
     FedAvg &45.3 &36.4 &68.8 &76.3 &\cellcolor[HTML]{efefef}56.7 \\
     FedAvg+GN &44.3 &31.1 &71.9 &74.6 &\cellcolor[HTML]{efefef}55.5 \\
     FedAvg+LN &34.4 &26.2 &59.4 &44.1 &\cellcolor[HTML]{efefef}41.0 \\
     FixBN &34.9 &33.8 &62.5 &78.0 &\cellcolor[HTML]{efefef}52.3 \\
     SiloBN &40.6 &29.3 &59.4 &81.4 &\cellcolor[HTML]{efefef}52.7 \\
     FedBN &57.3 &37.3 &90.6 &\textbf{89.8} &\cellcolor[HTML]{efefef}68.8 \\
     \textbf{FedWon} &\textbf{63.0} &\textbf{46.7} &\textbf{90.6} &86.4 &\cellcolor[HTML]{efefef}\textbf{71.7} \\
     \bottomrule
     \end{tabular}
     \endgroup
   \end{table}

\begin{figure}[t!]
  \centering
  \begin{subfigure}[t]{0.495\textwidth}
    \includegraphics[width=\textwidth]{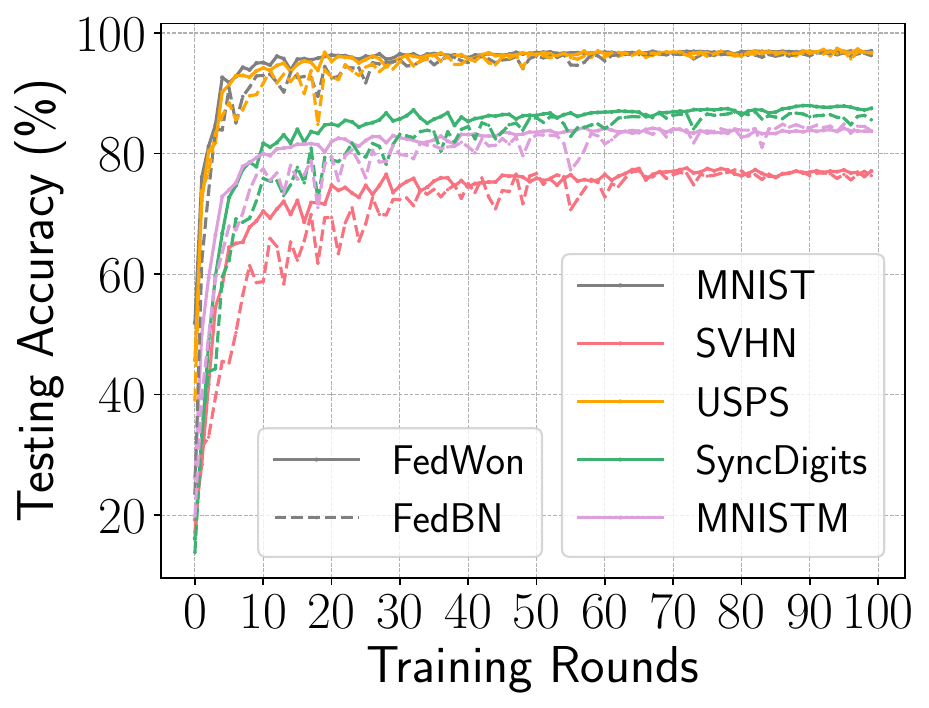}
    \caption{Cross-silo FL Training}
    \label{fig:digits-training-curve-b32}
  \end{subfigure}
  \hfill
  \begin{subfigure}[t]{0.495\textwidth}
     \includegraphics[width=\textwidth]{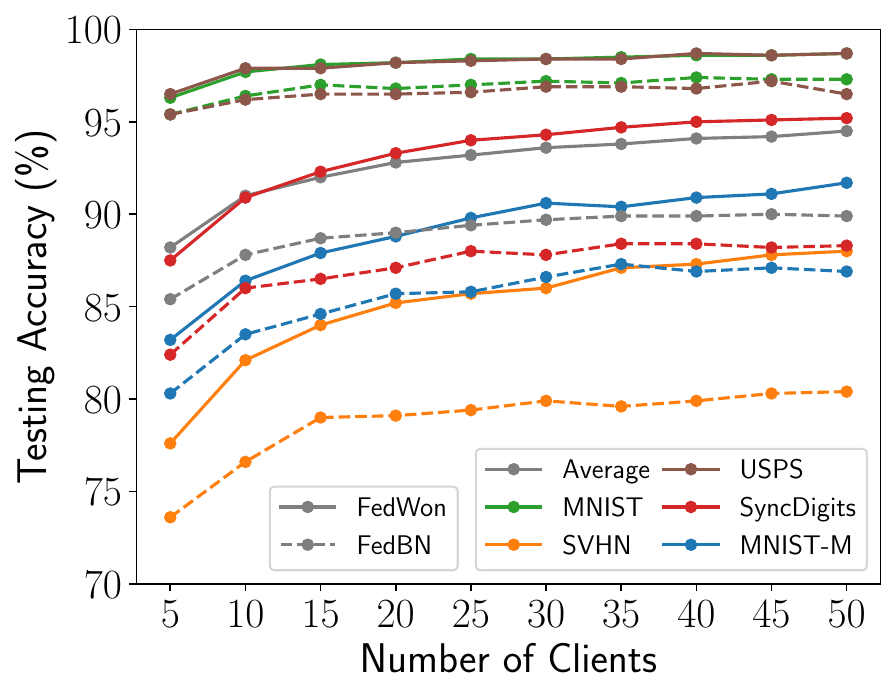}
     \caption{Domain Heterogeneity}
     \label{fig:domain-hetero}
 \end{subfigure}
 \caption{Testing accuracy (\%) comparison of FedBN and FedWon on Digits-Five dataset: (a) compares testing accuracy throughout training on cross-silo FL with total 5 clients (one client per domain) and batch size $B = 32$; (b) comparison of different degrees of domain heterogeneity.}
 \label{fig:digits-training-curve}
\end{figure}

  \begin{table}[t]\centering
    \caption{Performance comparison of FedBN and FedWon under different local epochs $E = \{1, 4, 8\}$ on Office-Caltech-10 dataset. FedWon maintains performance and consistently outperforms FedBN under different numbers of local epochs.}\label{tab:office-local-epoch}
    \begin{tabular}{c|c|cccccc}\toprule
    E &Methods &Amazon &Caltech &DSLR &Webcam &\cellcolor[HTML]{efefef}Average \\\midrule
    \multirow{2}{*}{1} &FedBN &67.2 &45.3 &85.4 &87.5 &\cellcolor[HTML]{efefef}71.4 \\
    &\textbf{FedWon} &\textbf{67.0} &\textbf{50.4} &\textbf{95.3} &\textbf{90.7} &\cellcolor[HTML]{efefef}\textbf{75.6} \\\midrule
    \multirow{2}{*}{4} &FedBN &66.7 &43.6 &84.4 &\textbf{89.8} &\cellcolor[HTML]{efefef}71.1 \\
    &\textbf{FedWon} &\textbf{68.8} &\textbf{51.1} &\textbf{93.8} &84.8 &\cellcolor[HTML]{efefef}\textbf{74.6} \\\midrule
    \multirow{2}{*}{8} &FedBN &64.6 &45.8 &87.5 &89.8 &\cellcolor[HTML]{efefef}71.9 \\
    &\textbf{FedWon} &\textbf{64.6} &\textbf{49.3} &\textbf{96.9} &\textbf{91.5} &\cellcolor[HTML]{efefef}\textbf{75.6} \\
    \bottomrule
    \end{tabular}
  \end{table}

\begin{table}[t]\centering
  \caption{Evaluation on cross-silo skewed label distribution using MobileNetV2 with 10 clients constructed by splitting the CIFAR-10 dataset with Dir (0.1).}\label{tab:cross-silo-skewed}
  \begin{tabular}{lcccccccc}\toprule
  Methods &FedAvg &FedAvg+GN &FedAvg+LN &SiloBN &FixBN &FedBN &FedWon \\\midrule
  Accuracy &66.95 &68.11 &71.65 &70.8 &66.22 &69.07 &76.45 \\
  \bottomrule
  \end{tabular}
  \end{table}

  \begin{table}[!htp]\centering
    \caption{Comparison of FedWon and FedAvg on a total of 1000 clients on Digits-Five dataset, with a selection of only 0.1 clients per round.}\label{tab:1000clients}
    \begin{tabular}{l|ccccc|c}\toprule
    Methods &MNIST &SVHN &USPS &SynthDigits &MNIST-M &Average \\ \midrule
    FedAvg &96.0 &71.2 &94.7 &82.9 &\textbf{82.8} &85.5 \\
    FedWon &\textbf{96.4} &\textbf{73.6} &\textbf{95.5} &\textbf{83.7} &81.9 &\textbf{86.2} \\
    \bottomrule
    \end{tabular}
    \end{table}

  \begin{table}[!htp]\centering
    \caption{Comparison of FedWon and PartialFed on Office-Caltech-10 dataset.}\label{tab:comp-partialfed}
    \begin{tabular}{l|cccc|c}\toprule
    Methods &Amazon &Caltech &DSLR &Webcam &Average \\\midrule
    PartialFed-Fix &58.3 &44.9 &88.1 &\textbf{91.2} &70.6 \\
    PartialFed-Adaptive &63.4 &45.4 &85.6 &90.5 &71.3 \\
    FedWon (Ours) &\textbf{67.0} &\textbf{50.4} &\textbf{95.3} &90.7 &\textbf{75.6} \\
    \bottomrule
    \end{tabular}
  \end{table}

  \begin{table}[t!]\centering
    \caption{Ablation studies on the impact of WSConv and AGC}\label{tab:ablation}
    \begin{tabular}{c|c|c|cccccc}\toprule
    Batch Size &WSConv &AGC &Amazon &Caltech &DSLR &Webcam &\cellcolor[HTML]{efefef}Average \\\midrule
    \multirow{4}{*}{32} &$\checkmark$ &$\checkmark$ &63.7 &51.0 &96.3 &91.2 &\cellcolor[HTML]{efefef}75.6 \\
    &$\checkmark$ & &65.1 &52.0 &90.6 &89.8 &\cellcolor[HTML]{efefef}74.4 \\
    & &$\checkmark$ &46.4 &37.3 &68.8 &71.2 &\cellcolor[HTML]{efefef}55.9 \\
    & & &27.1 &19.6 &37.5 &28.8 &\cellcolor[HTML]{efefef}28.2 \\\midrule
    \multirow{4}{*}{2} &$\checkmark$ &$\checkmark$ &65.1 &51.1 &93.8 &86.4 &\cellcolor[HTML]{efefef}74.1 \\
    &$\checkmark$ & &67.2 &55.6 &96.9 &93.2 &\cellcolor[HTML]{efefef}78.2 \\
    & &$\checkmark$ &53.1 &38.7 &87.5 &78.0 &\cellcolor[HTML]{efefef}64.3 \\
    & & &54.7 &44.0 &84.4 &78.0 &\cellcolor[HTML]{efefef}65.3 \\
    \bottomrule
    \end{tabular}
  \end{table}

\begin{figure}[t]
  \centering
  \begin{subfigure}[t]{0.48\textwidth}
      \includegraphics[width=\textwidth]{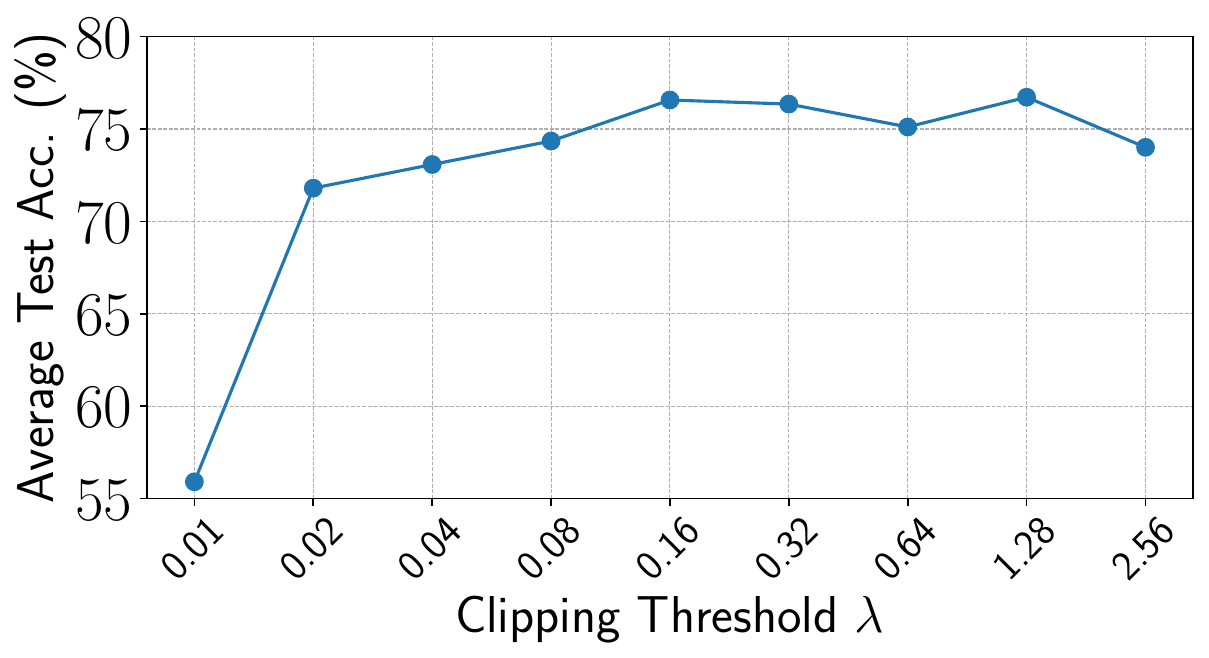}
      \caption{Batch Size $B = 32$}
      \label{fig:clipping-b32}
  \end{subfigure}
  \hfill
  \begin{subfigure}[t]{0.48\textwidth}
     \includegraphics[width=\textwidth]{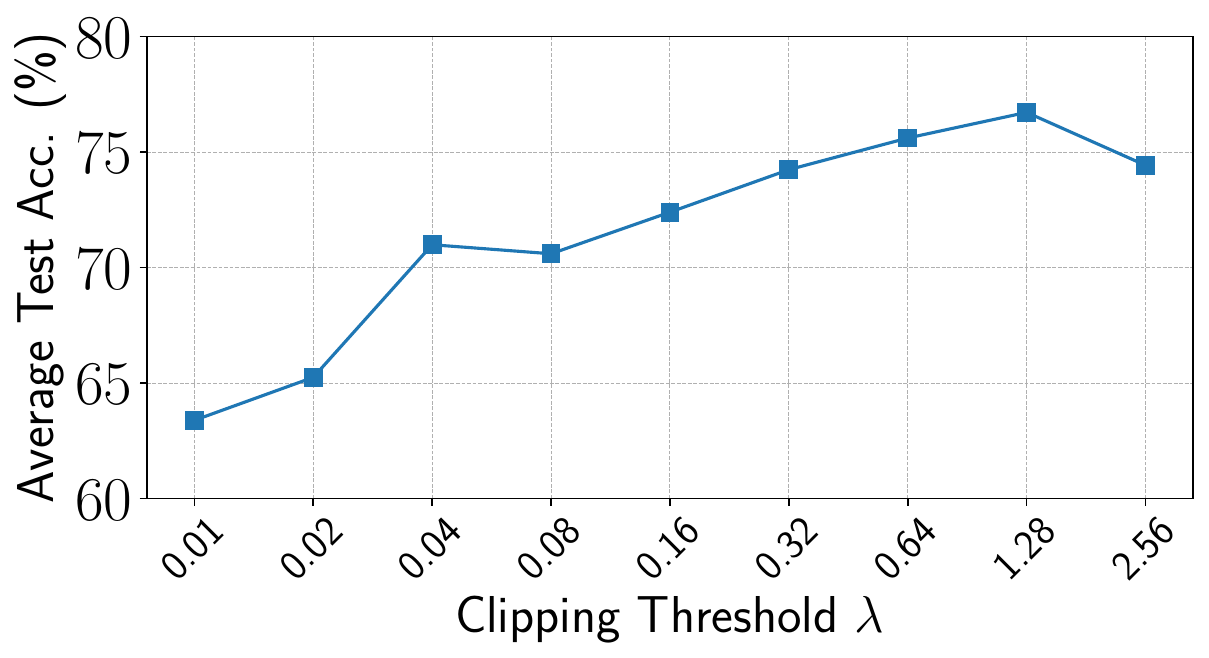}
     \caption{Batch Size $B = 2$}
     \label{fig:clipping-b2}
 \end{subfigure}
 \caption{Impact of clipping threshold $\lambda$ of adaptive gradient clipping (AGC) on Office-Caltech-10 dataset, using different batch sizes $B$.}
 \label{fig:clipping-threshold}
\end{figure}

\subsection{Additional Analysis}

\textbf{Domain Generalization Capability.} We expand our analysis to investigate the domain adaptation and generalization capabilities of FedWon. Our experiments are conducted on the Office-Caltech-10 dataset, where we employ Amazon (A), Caltech (C), and DSLR (D) as the seen domains during training, while WebCam (W) is exclusively reserved as an unseen domain only for evaluation, specifically for zero-shot evaluation. We use the client local models to evaluate the seen domains and use the server global model to test on the unseen domain; while to be fair for FedBN, we employ a global model with averaged BN layer parameters from the seen domains. Table \ref{tab:domain-generalization} presents compelling evidence that FedWon not only excels in performance on the seen domains but also exhibits the most robust generalization capabilities on the unseen domains. These results demonstrate an additional advantage of FedWon, highlighting its capability for domain generalization.

 \textbf{Evaluation on Alternative Backbones.} In addition to evaluating the effectiveness of FedWon using AlexNet \citep{krizhevsky2017alexnet} on the Office-Caltech-10 dataset, 
 Table \ref{tab:office-resnet} also compares testing accuracy on a common backbone, ResNet-20 \citep{he2016resnet}. Interestingly, replacing BN with GN or LN is not as effective on ResNet-20 as on AlexNet. FedAvg+GN and FedAvg+LN only achieve similar or even worse performance than FedAvg. FedBN \citep{li2021fedbn}, instead, achieves better performance than the other existing methods. Nevertheless, our proposed FedWon consistently outperforms the state-of-the-art methods even with ResNet-20 as the backbone.

\textbf{Analysis on Different Degrees of Domain Heterogeneity.} We evaluate the performance of the proposed FedWon under different degrees of domain heterogeneity. To simulate varying degrees of domain heterogeneity, we follow the approach taken by FedBN \citep{li2021fedbn} and create different numbers of clients with the same domain on the Digits-Five dataset. We start with 5 clients, each containing data from one domain, and then add 5 clients at a time, with each new client containing one of the Digits-Five datasets, respectively. We evaluate the performance of the algorithms for different numbers of clients 
from \(N = \{5, 10, 15, ..., 50\}\). 
More clients represent less heterogeneity as more clients have overlapping domains of data.
Figure \ref{fig:domain-hetero} compares the performance of FedWon and FedBN under these settings. The results show that the performances of both FedWon and FedBN increase as the degree of heterogeneity decreases. FedBN outperforms FedAvg in all the settings as evidenced in \cite{li2021fedbn}. However, our proposed FedWon achieves even better performance than FedBN on all domains and all levels of heterogeneity.

\textbf{Testing Accuracy Changes Throughout Training.} Figure \ref{fig:digits-training-curve} illustrates the changes in testing accuracy throughout the training process on the Digits-Five dataset. Specifically, Figure \ref{fig:digits-training-curve-b32} compares the performance of FedWon and FedBN in a cross-silo FL involving a total of 5 clients (one client per domain) and a batch size of $B = 32$. FedWon outperforms FedBN in certain domains or demonstrates similar performance in others. Notably, FedWon achieves better performance in the early stage of training -- FedWon exhibits faster convergence, achieving a satisfactory level of accuracy more quickly than FedBN. These results complement the results in Figure \ref{fig:mnist} (right) in the main manuscript that compares FedWon and FedAvg in a cross-device FL scenario.

\textbf{Impact of Local Epochs.} Table \ref{tab:office-local-epoch} compares the performance of our proposed FedWon and FedBN \citep{li2021fedbn} under different local epochs $E = \{1, 4, 8\}$ on Office-Caltech-10 dataset. FedWon maintains performance and consistently outperforms FedBN under different numbers of local epochs. We run these experiments with batch size $B = 32$ and the learning rate the same as the ones in Table \ref{tab:main-lr} on the Office-Caltech-10 dataset.

\textbf{Evaluation on Cross-silo FL for Skewed Label Distribution.} Table \ref{tab:cross-silo-skewed} compares different algorithms in cross-silo FL for skewed label distribution on the CIFAR-10 dataset, which complements cross-device FL experiments in Figure \ref{fig:skewed-label}. FedWon also consistently outperforms all other methods in cross-silo FL. We run experiments with 10 clients under Dir(0.1) of non-i.i.d data, batch size $B = 64$, and local epoch $E = 5$ for 200 rounds. 

\textbf{Evaluation on Cross-device FL of 1000 clients.} Table \ref{tab:1000clients} compares FedWon and FedAvg on a total of 1000 clients on Digits-Five dataset, with a selection of only 0.1 clients per round. FedWon also generally outperforms FedAvg under this setting. These experiments are run with batch size B = 2 and learning rate of 0.02. 

\textbf{Addtional Comparison with PartialFed.} Table \ref{tab:comp-partialfed} further compares FedWon with two variant implementations of PartialFed \citep{sun2021partialfed} on Office-Caltech-10 dataset. FedWon generally achieves superior performance to PartialFed, especially on the average testing accuracy.

\subsection{Additional Ablation Studies}

\begin{table}[t!]\centering
  \caption{Evaluation on the impact of using AGC optimizer on different algorithms. AGC also benefits other methods, while our proposed FedWon achieves the best overall performance.}\label{tab:agc}
  \begin{tabular}{lccccccc}\toprule
  Methods &AGC &Amazon &Caltech &DSLR &Webcam &\cellcolor[HTML]{efefef}Average \\\midrule
  FedAvg & &61.8 &44.9 &77.1 &81.4 &\cellcolor[HTML]{efefef}66.3 \\
  FedAvg & \checkmark &62.5 &45.3 &75.0 &84.8 &\cellcolor[HTML]{efefef}66.9 \\\midrule
  FedAvg+GN & &60.8 &50.8 &88.5 &83.6 &\cellcolor[HTML]{efefef}70.9 \\
  FedAvg+GN & \checkmark &64.1 &48.0 &90.6 &88.1 &\cellcolor[HTML]{efefef}72.7 \\\midrule
  FedAvg+LN & &55.0 &41.3 &79.2 &71.8 &\cellcolor[HTML]{efefef}61.8 \\
  FedAvg+LN & \checkmark &59.4 &42.2 &84.4 &79.7 &\cellcolor[HTML]{efefef}66.4 \\\midrule
  FixBN & &59.2 &44.0 &79.2 &79.6 &\cellcolor[HTML]{efefef}65.5 \\
  FixBN & \checkmark &58.9 &43.1 &75.0 &88.1 &\cellcolor[HTML]{efefef}66.3 \\\midrule
  SiloBN & &60.8 &44.4 &76.0 &81.9 &\cellcolor[HTML]{efefef}65.8 \\
  SiloBN & \checkmark &59.4 &44.4 &78.1 &83.0 &\cellcolor[HTML]{efefef}66.2 \\\midrule
  FedBN & &\textbf{67.2} &45.3 &85.4 &87.5 &\cellcolor[HTML]{efefef}71.4 \\
  FedBN & \checkmark &\textbf{70.3} &45.3 &87.5 &88.1 &\cellcolor[HTML]{efefef}72.8 \\\midrule
  FedWon (Ours) & \checkmark &67.0 &\textbf{50.4} &\textbf{95.3} &\textbf{90.7} &\cellcolor[HTML]{efefef}\textbf{75.6} \\
  \bottomrule
  \end{tabular}
\end{table}

\textbf{Impact of WSConv and AGC.} We analyze the impact of WSConv and AGC, which supplements the ablation study presented in the main manuscript. Table \ref{tab:ablation} shows the impact of these two components with batch size $B = 32$ and small batch size $B = 2$ on the Office-Caltech-10 dataset. After removing the normalizations, using WSConv significantly improves the performance on both batch sizes. AGC, however, shows a positive impact only with batch size $B = 32$, as it is specifically designed for larger batch sizes. Consequently, we do not adopt AGC in the experiments with small batch sizes ($B =\{1, 2, 4\}$). We run these experiments with learning rate $\eta = 0.08$ for $B = 32$ and $\eta = 0.01$ for $B = 2$.

\textbf{Impact of Clipping Threshold $\lambda$ for AGC.} We further extend to evaluate the impact of clipping threshold $\lambda$ under batch sizes $B = 2$ and $B = 32$. Figure \ref{fig:clipping-threshold} shows the average testing accuracy on the Office-Caltech-10 dataset using different clipping thresholds $\lambda = \{0.01, 0.02, 0.04, 0.08, 0.16, 0.32, 0.64, 1.28, 2.56\}$. When the batch size $B = 32$, the performance is rather insensitive to different values of $\lambda$ when it is not too small (larger than 0.08). When the batch size $B = 2$, the best clipping threshold is $\lambda = 1.28$ and the performance is sensitive to different values. Consistent with the finding in Table \ref{tab:ablation}, we recommend avoiding using AGC when the batch size is small. These results provide insights into selecting an appropriate clipping threshold for multi-domain FL.

\textbf{Impact of AGC on Other Algorithms.} Table \ref{tab:agc} compares the results with and without AGC with the same learning rate for different methods. AGC also benefits other methods, while our proposed FedWon achieves the best overall performance.

\subsection{Complementary Experiments}

\begin{figure}[t!]
  \centering
  \begin{subfigure}{\textwidth}
  \includegraphics[width=\textwidth]{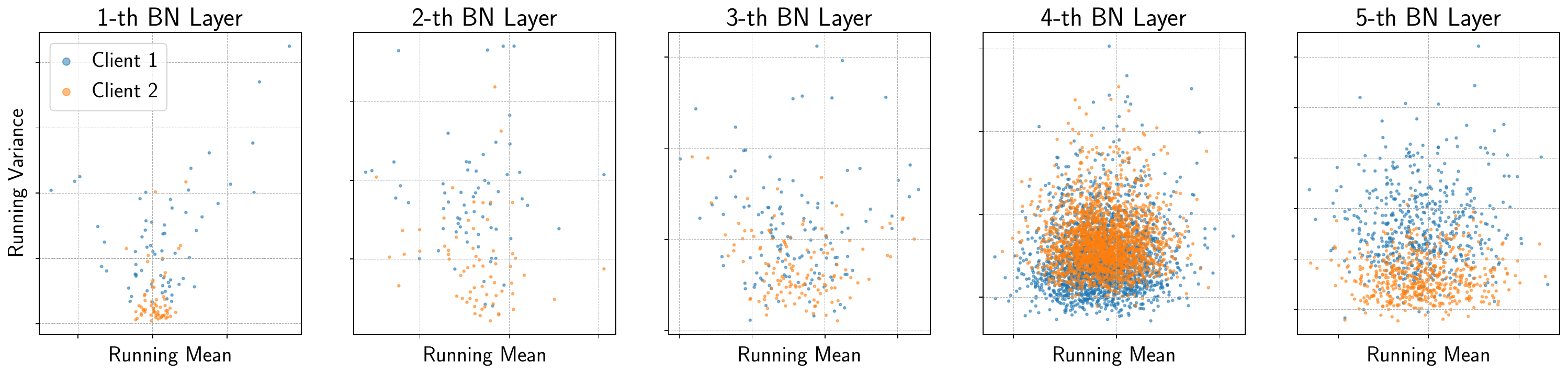}
  \caption{BN statistics of models in two clients.}
  \end{subfigure}
  \begin{subfigure}{\textwidth}
    \includegraphics[width=\textwidth]{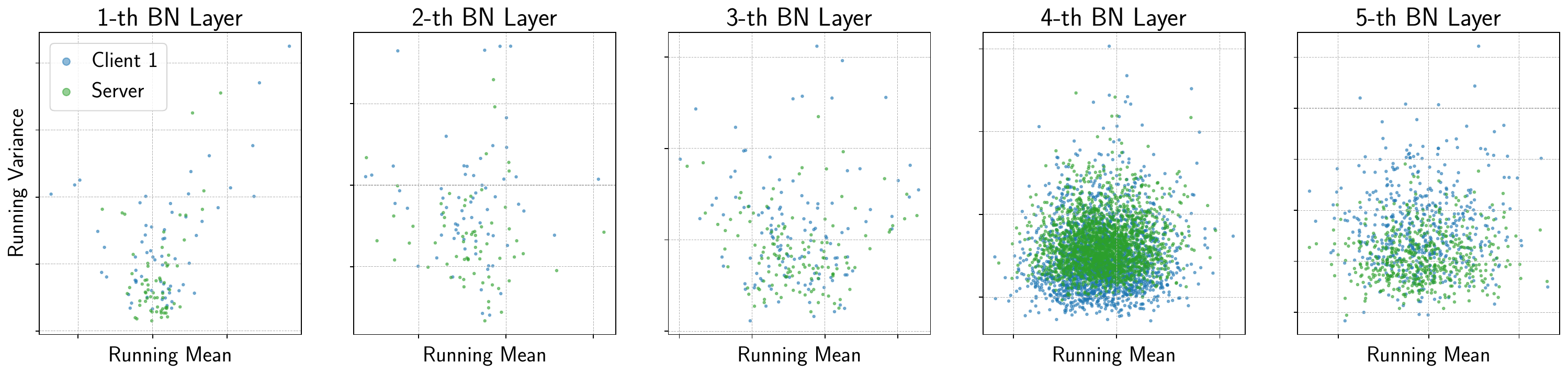}
    \caption{BN statistics of models in a client (Client 1) and the server.}
  \end{subfigure}
  \begin{subfigure}{\textwidth}
      \includegraphics[width=\textwidth]{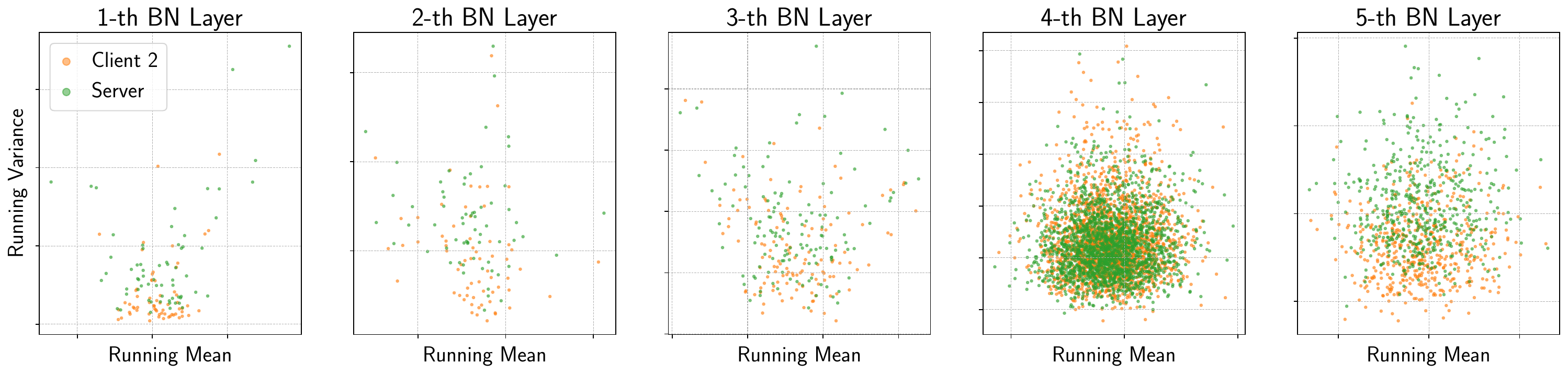}
      \caption{BN statistics of models in a client (Client 2) and the server.}
    \end{subfigure}    
 \caption{BN statistics of all layers in a 6-layer CNN.}
 \label{fig:bn-stats-all}
\end{figure}

\begin{figure}[t!]

  \hspace*{\fill}%
  \begin{minipage}[t]{0.49\textwidth}
  \centering
  \vspace{0pt}
  \begin{subfigure}{\textwidth}
      \includegraphics[width=\textwidth]{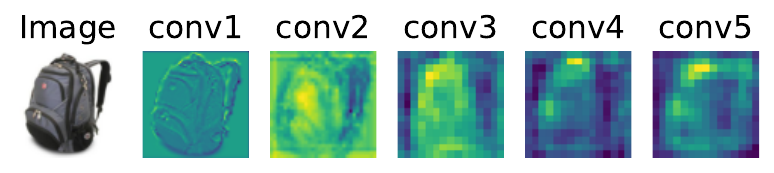}
      \caption{Amazon - FedAvg}
      \label{fig:amazon-fedavg}
  \end{subfigure}

  \begin{subfigure}{\textwidth}
    \includegraphics[width=\textwidth]{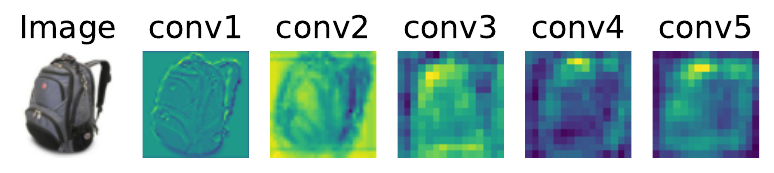}
    \caption{Amazon - FedBN}
    \label{fig:amazon-fedbn}
  \end{subfigure}

  \begin{subfigure}{\textwidth}
    \includegraphics[width=\textwidth]{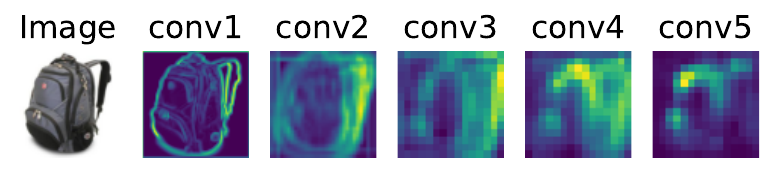}
    \caption{Amazon - FedWon}
    \label{fig:amazon-fedwon}
  \end{subfigure}

  \begin{subfigure}{\textwidth}
    \includegraphics[width=\textwidth]{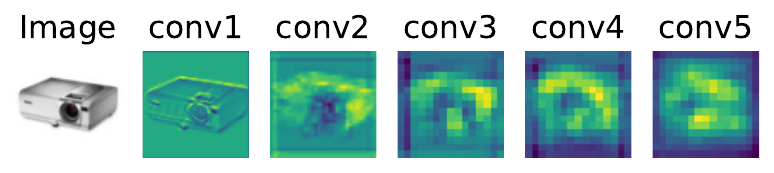}
    \caption{Caltech - FedAvg}
    \label{fig:Caltech-fedavg}
  \end{subfigure}

  \begin{subfigure}{\textwidth}
    \includegraphics[width=\textwidth]{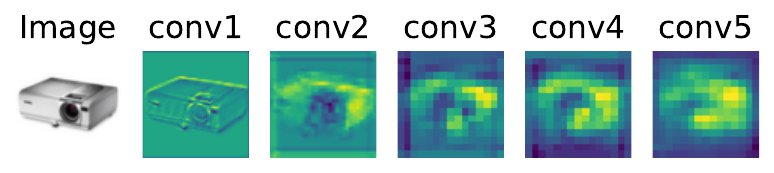}
    \caption{Caltech - FedBN}
    \label{fig:Caltech-fedbn}
  \end{subfigure}

  \begin{subfigure}{\textwidth}
    \includegraphics[width=\textwidth]{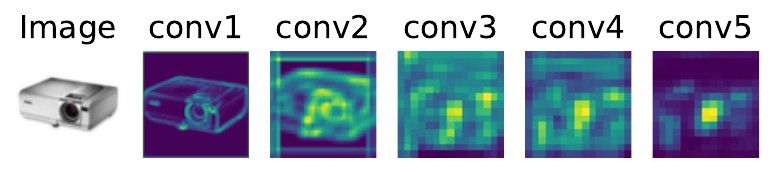}
    \caption{Caltech - FedWon}
    \label{fig:Caltech-fedwon}
  \end{subfigure}
  \end{minipage}%
  \hfill
  \begin{minipage}[t]{0.49\textwidth}
  \centering
  \vspace{0pt}
  \begin{subfigure}{\textwidth}
    \includegraphics[width=\textwidth]{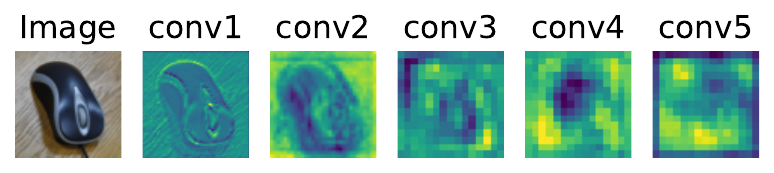}
    \caption{DSLR - FedAvg}
    \label{fig:DSLR-fedavg}
  \end{subfigure}

  \begin{subfigure}{\textwidth}
    \includegraphics[width=\textwidth]{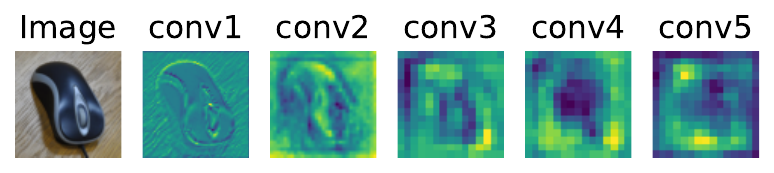}
    \caption{DSLR - FedBN}
    \label{fig:DSLR-fedbn}
  \end{subfigure}

  \begin{subfigure}{\textwidth}
  \includegraphics[width=\textwidth]{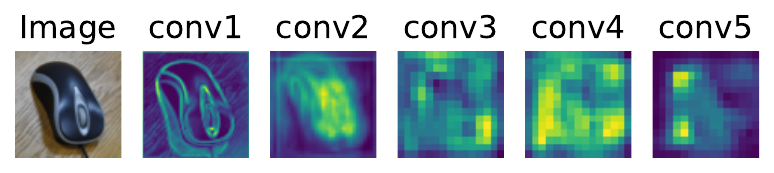}
  \caption{DSLR - FedWon}
  \label{fig:DSLR-fedwon}
  \end{subfigure}

  \begin{subfigure}{\textwidth}
    \includegraphics[width=\textwidth]{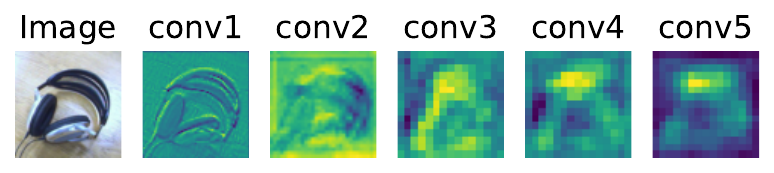}
    \caption{WebCam - FedAvg}
    \label{fig:WebCam-fedavg}
  \end{subfigure}

  \begin{subfigure}{\textwidth}
    \includegraphics[width=\textwidth]{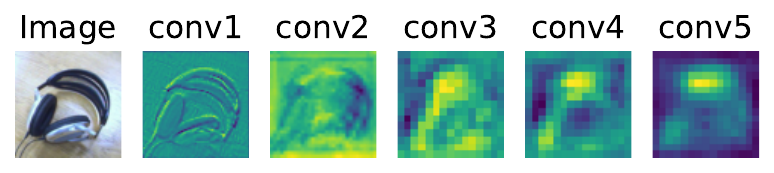}
    \caption{WebCam - FedBN}
    \label{fig:WebCam-fedbn}
  \end{subfigure}

  \begin{subfigure}{\textwidth}
    \includegraphics[width=\textwidth]{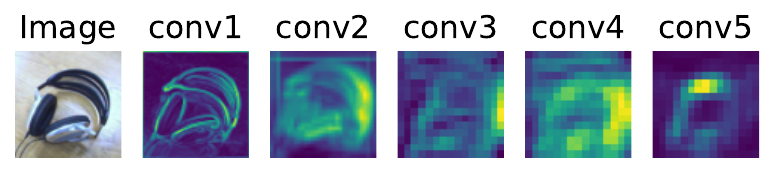}
    \caption{WebCam - FedWon}
    \label{fig:WebCam-fedwon}
  \end{subfigure}
  \end{minipage}%

  \hspace*{\fill}
  \caption{Visualization of feature maps of FedAvg, FedBN, and our proposed FedWon on the Office-Caltech-10 dataset, which contain four domains: Amazon, Caltech, DSLR, and WebCam.}
  \label{fig:visualization-all-conv}
\end{figure} 

\textbf{Visualization of BN Statistics.} Figure \ref{fig:bn-stats-all} visualizes the running mean and variance of BN layers of the 6-layer CNN. It complements Figure \ref{fig:bn-stats} in the main manuscript and shows the discrepancies of BN statistics between clients and between a client and the server in all BN layers.

\textbf{Visualization of Feature Maps.} Figure \ref{fig:visualization-all-conv} presents the visualization of feature maps obtained through three methods: FedAvg, FedBN, and our proposed FedWon. These feature maps are the output of each convolution layer in AlexNet on the Office-Caltech-10 dataset, which encompasses data from four distinct domains, namely Amazon, Caltech, DSLR, and WebCam. FedWon exhibits significantly enhanced feature maps on the object of interest compared to those produced by the FedAvg and FedBN. 

\begin{figure}[t!]
  \centering
    \includegraphics[width=0.6\textwidth]{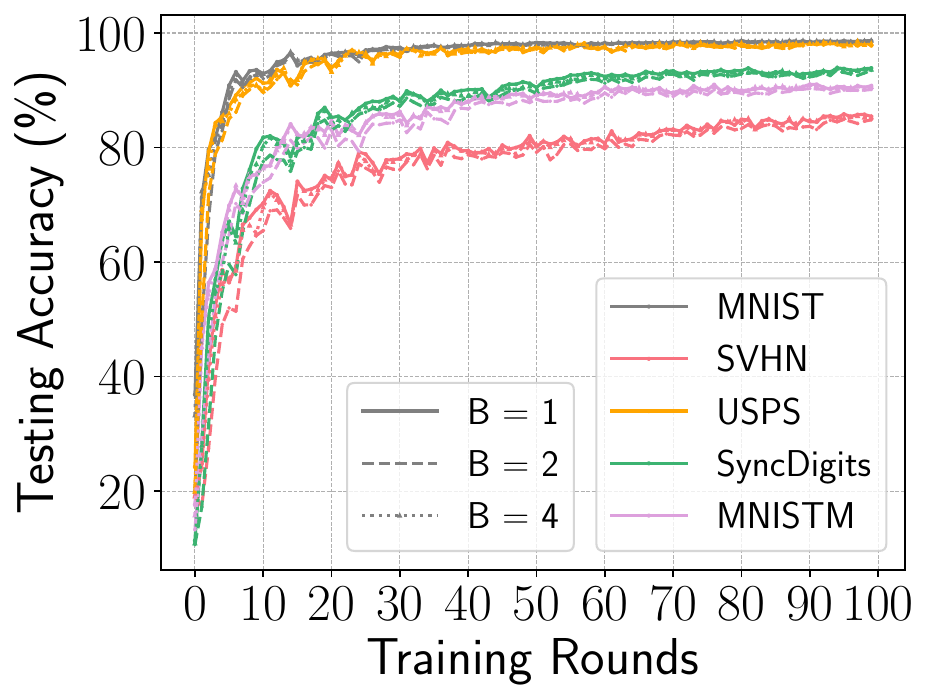}
 \caption{Testing accuracy (\%) comparison of different batch sizes $B = \{1, 2, 4\}$ using FedWon on Digits-Five dataset with 10 randomly selected clients out of 100 clients.}
 \label{fig:digits-training-curve-batches}
\end{figure}

\textbf{Effectiveness on Small Size.} Table \ref{tab:small-batch-office-b4} compares performance of our proposed FedWon with existing methods on Office-Caltech-10 dataset with batch size $B = 4$. FedWon achieves the best performance also in this setting, complementing the experiments of batch size $B = \{1, 2\}$ in Table \ref{tab:small-batch-office} in the main manuscript. Additionally, Figure \ref{fig:digits-training-curve-batches} compares the testing accuracy over the course of training of FedWon with batch sizes $B = \{1, 2, 4\}$ on Digits-Five dataset. Different batch sizes tend to have a similar trend of convergence.

\textbf{Effectiveness on Selection of Clients.} Table \ref{tab:client-fraction-04} compares the performance of FedAvg and FedWon on cross-device FL on the Digits-Five dataset with a fraction $C = 0.4$ of clients out of a total of 100 clients to participate in training each round. FedWon achieves superior performance also in this setting, complementing the experiments of $C = \{0.1, 0.2\}$ in Table \ref{tab:client-fraction} in the main manuscript.

\textbf{Comparison of Methods with Variances.} Table \ref{tab:cross-silo} presents testing accuracy comparison of different methods on three datasets with mean (standard deviation) of three runs of experiments. It complements the results in Table \ref{tab:cross-silo-mean} in the main manuscript.

  \begin{table}[t!]
    \centering
     \caption{Performance comparison using small batch sizes \(B = 4 \) on Office-Caltech-10 dataset. Our proposed FedWon achieves outstanding performance compared to existing methods.}\label{tab:small-batch-office-b4}
     \begingroup
     \setlength{\tabcolsep}{0.47em}    
       \begin{tabular}{l|l|ccccc}\toprule
       B &Methods &Amazon &Caltech &DSLR &WebCam \\\midrule
     \multirow{7}{*}{4} &FedAvg &65.6 &46.7 &78.1 &88.1 \\
     &FedAvg+GN &60.9 &52.0 &84.4 &89.8 \\
     &FedAvg+LN &54.2 &44.9 &78.1 &72.9 \\
     &FixBN &66.2 &50.2 &78.1 &91.5 \\
     &SiloBN &63.5 &48.9 &78.1 &88.1 \\
     &FedBN &67.2 &50.7 &90.6 &91.5 \\
     &\textbf{FedWon} &\textbf{68.8} &\textbf{54.2} &\textbf{96.9} &\textbf{91.5} \\
     \bottomrule
     \end{tabular}
     \endgroup
  \end{table}

\begingroup
\setlength{\tabcolsep}{0.44em}
\begin{table}[t]\centering
  \caption{Testing accuracy comparison on randomly selecting a fraction \(C = 0.4 \) out of a total of 100 clients for training each round with batch size $B = 4$. FedWon consistently outperforms FedAvg on Digits-Five dataset. We report the mean (standard deviation) of three runs of experiments.}\label{tab:client-fraction-04}
  \begin{tabular}{c|l|cccccc}\toprule
  C &Method &MNIST &SVHN &USPS &SynthDigits &MNIST-M &\cellcolor[HTML]{efefef}Average \\\midrule
  \multirow{2}{*}{0.4} &FedAvg &98.1 {\footnotesize (0.0)} &80.5 {\footnotesize (0.1)} &97.0 {\footnotesize (0.2)} &91.4 {\footnotesize (0.2)} &89.4 {\footnotesize (0.1)} &\cellcolor[HTML]{efefef}91.3 {\footnotesize (0.0)} \\
  &\textbf{FedWon {\footnotesize (Ours)} }&\textbf{98.8 {\footnotesize (0.0)}} &\textbf{86.4 {\footnotesize (0.2)}} &\textbf{98.4 {\footnotesize (0.2)}} &\textbf{94.2 {\footnotesize (0.2)}} &\textbf{91.0 {\footnotesize (0.3)}} &\cellcolor[HTML]{efefef}\textbf{93.7 {\footnotesize (0.0)}} \\
  \bottomrule
  \end{tabular}
\end{table}
\endgroup

\begingroup
\setlength{\tabcolsep}{0.32em}
\begin{table}[t]\centering
  \caption{Testing accuracy (\%) comparison of different methods on three datasets. Our proposed FedWon outperforms existing methods in most of the domains. FedWon achieves the best average testing accuracy in all datasets. We report the mean (standard deviation) of three runs of experiments.}\label{tab:cross-silo}
  \begin{tabular}{l|l|cccccccc|c}\toprule
  &Domains &Standalone &FedAvg &FedProx &+GN $^a$ &+LN $^b$ &SiloBN &FixBN &FedBN &Ours \\\midrule
  \multirow{12}{*}{\rotatebox[origin=c]{90}{Digit-Five}} &\multirow{2}{*}{MNIST} &94.4 &96.2 &96.4 &96.4 &96.4 &96.2 &96.3 &96.5 &\textbf{96.8} \\
  & &{\footnotesize (0.2)} &{\footnotesize (0.2)} &{\footnotesize (0.0)} &{\footnotesize (0.1)} &{\footnotesize (0.1)} &{\footnotesize (0.0)} &{\footnotesize (0.1)} &{\footnotesize (0.1)} &\textbf{{\footnotesize (0.2)}} \\
  &\multirow{2}{*}{SVHN} &67.1 &71.6 &71.0 &76.9 &75.2 &71.3 &71.3 &77.3 &\textbf{77.4} \\
  & &{\footnotesize (0.7)} &{\footnotesize (0.5)} &{\footnotesize (0.8)} &{\footnotesize (0.1)} &{\footnotesize (0.4)} &{\footnotesize (1.0)} &{\footnotesize (0.9)} &{\footnotesize (0.4)} &\textbf{{\footnotesize (0.1)}} \\
  &\multirow{2}{*}{USPS} &95.4 &96.3 &96.1 &96.6 &96.4 &96.0 &96.1 &96.9 &\textbf{97.0} \\
  & &{\footnotesize (0.1)} &{\footnotesize (0.3)} &{\footnotesize (0.1)} &{\footnotesize (0.2)} &{\footnotesize (0.4)} &{\footnotesize (0.2)} &{\footnotesize (0.2)} &{\footnotesize (0.2)} &\textbf{{\footnotesize (0.1)}} \\
  &\multirow{2}{*}{SynthDigits} &80.3 &86.0 &85.9 &86.6 &85.6 &86.0 &85.8 &86.8 &\textbf{87.6} \\
  & &{\footnotesize (0.8)} &{\footnotesize (0.3)} &{\footnotesize (0.2)} &{\footnotesize (0.1)} &{\footnotesize (0.3)} &{\footnotesize (0.3)} &{\footnotesize (0.1)} &{\footnotesize (0.3)} &\textbf{{\footnotesize (0.2)}} \\
  &\multirow{2}{*}{MNIST-M} &77.0 &82.5 &83.1 &83.7 &82.2 &83.1 &83.0 &\textbf{84.6} &84.0 \\
  & &{\footnotesize (0.9)} &{\footnotesize (0.1)} &{\footnotesize (0.2)} &{\footnotesize (0.5)} &{\footnotesize (0.3)} &{\footnotesize (0.4)} &{\footnotesize (0.8)} &\textbf{{\footnotesize (0.2)}} &{\footnotesize (0.2)} \\
  &\cellcolor[HTML]{efefef} &\cellcolor[HTML]{efefef}83.1 &\cellcolor[HTML]{efefef}86.5 &\cellcolor[HTML]{efefef}86.5 &\cellcolor[HTML]{efefef}88.0 &\cellcolor[HTML]{efefef}87.1 &\cellcolor[HTML]{efefef}86.5 &\cellcolor[HTML]{efefef}86.5 &\cellcolor[HTML]{efefef}88.4 &\cellcolor[HTML]{efefef}\textbf{88.5} \\
  &\multirow{-2}{*}{\cellcolor[HTML]{efefef}Average} &\cellcolor[HTML]{efefef}{\footnotesize (0.4)} &\cellcolor[HTML]{efefef}{\footnotesize (0.1)} &\cellcolor[HTML]{efefef}{\footnotesize (0.1)} &\cellcolor[HTML]{efefef}{\footnotesize (0.1)} &\cellcolor[HTML]{efefef}{\footnotesize (0.0)} &\cellcolor[HTML]{efefef}{\footnotesize (0.3)} &\cellcolor[HTML]{efefef}{\footnotesize (0.0)} &\cellcolor[HTML]{efefef}{\footnotesize (0.1)} &\cellcolor[HTML]{efefef}\textbf{{\footnotesize (0.1)}} \\
  \midrule
  \multirow{10}{*}{\rotatebox[origin=c]{90}{Office-Caltech-10}} &\multirow{2}{*}{Amazon} &54.5 &61.8 &59.9 &60.8 &55.0 &60.8 &59.2 &\textbf{67.2} &67.0 \\
  & &{\footnotesize (1.8)} &{\footnotesize (1.2)} &{\footnotesize (0.5)} &{\footnotesize (1.8)} &{\footnotesize (0.3)} &{\footnotesize (1.3)} &{\footnotesize (1.8)} &\textbf{{\footnotesize (0.9)}} &{\footnotesize (0.7)} \\
  &\multirow{2}{*}{Caltech} &40.2 &44.9 &44.0 &50.8 &41.3 &44.4 &44.0 &45.3 &\textbf{50.4} \\
  & &{\footnotesize (0.7)} &{\footnotesize (1.2)} &{\footnotesize (1.9)} &{\footnotesize (3.3)} &{\footnotesize (1.2)} &{\footnotesize (1.2)} &{\footnotesize (0.8)} &{\footnotesize (1.3)} &\textbf{{\footnotesize (2.8)}} \\
  &\multirow{2}{*}{DSLR} &81.3 &77.1 &76.0 &88.5 &79.2 &76.0 &79.2 &85.4 &\textbf{95.3} \\
  & &{\footnotesize (0.0)} &{\footnotesize (1.8)} &{\footnotesize (1.8)} &{\footnotesize (1.8)} &{\footnotesize (1.8)} &{\footnotesize (1.8)} &{\footnotesize (1.8)} &{\footnotesize (1.8)} &\textbf{{\footnotesize (2.2)}} \\
  &\multirow{2}{*}{Webcam} &89.3 &81.4 &80.8 &83.6 &71.8 &81.9 &79.6 &87.5 &\textbf{90.7} \\
  & &{\footnotesize (1.0)} &{\footnotesize (1.7)} &{\footnotesize (2.6)} &{\footnotesize (5.2)} &{\footnotesize (2.0)} &{\footnotesize (2.0)} &{\footnotesize (2.9)} &{\footnotesize (1.0)} &\textbf{{\footnotesize (1.2)}} \\
  &\cellcolor[HTML]{efefef} &\cellcolor[HTML]{efefef}66.3 &\cellcolor[HTML]{efefef}66.3 &\cellcolor[HTML]{efefef}65.2 &\cellcolor[HTML]{efefef}70.9 &\cellcolor[HTML]{efefef}61.8 &\cellcolor[HTML]{efefef}65.8 &\cellcolor[HTML]{efefef}65.5 &\cellcolor[HTML]{efefef}71.4 &\cellcolor[HTML]{efefef}\textbf{75.6} \\
  &\multirow{-2}{*}{\cellcolor[HTML]{efefef}Average} &\cellcolor[HTML]{efefef}{\footnotesize (0.4)} &\cellcolor[HTML]{efefef}{\footnotesize (0.7)} &\cellcolor[HTML]{efefef}{\footnotesize (1.0)} &\cellcolor[HTML]{efefef}{\footnotesize (2.5)} &\cellcolor[HTML]{efefef}{\footnotesize (0.7)} &\cellcolor[HTML]{efefef}{\footnotesize (0.2)} &\cellcolor[HTML]{efefef}{\footnotesize (0.8)} &\cellcolor[HTML]{efefef}{\footnotesize (1.0)} &\cellcolor[HTML]{efefef}\textbf{{\footnotesize (1.4)}} \\
  \midrule
  \multirow{14}{*}{\rotatebox[origin=c]{90}{DomainNet}} &\multirow{2}{*}{Clipart} &42.7 &48.9 &51.1 &45.4 &42.7 &51.8 &49.2 &49.9 &\textbf{57.2} \\
  & &{\footnotesize (2.7)} &{\footnotesize (2.0)} &{\footnotesize (0.8)} &{\footnotesize (0.5)} &{\footnotesize (0.7)} &{\footnotesize (1.0)} &{\footnotesize (1.8)} &{\footnotesize (0.5)} &\textbf{{\footnotesize (0.5)}} \\
  &\multirow{2}{*}{Infograph} &24.0 &26.5 &24.1 &21.1 &23.6 &25.0 &24.5 &28.1 &\textbf{28.1} \\
  & &{\footnotesize (1.6)} &{\footnotesize (2.5)} &{\footnotesize (1.6)} &{\footnotesize (1.1)} &{\footnotesize (1.2)} &{\footnotesize (2.1)} &{\footnotesize (0.9)} &{\footnotesize (0.8)} &\textbf{{\footnotesize (0.2)}} \\
  &\multirow{2}{*}{Painting} &34.2 &37.7 &37.3 &35.4 &35.3 &36.4 &38.2 &40.4 &\textbf{43.7} \\
  & &{\footnotesize (1.6)} &{\footnotesize (3.3)} &{\footnotesize (2.0)} &{\footnotesize (2.0)} &{\footnotesize (0.6)} &{\footnotesize (1.9)} &{\footnotesize (0.7)} &{\footnotesize (0.7)} &\textbf{{\footnotesize (1.2)}} \\
  &\multirow{2}{*}{Quickdraw} &\textbf{71.6} &44.5 &46.1 &57.2 &46.0 &45.9 &46.3 &69.0 &69.2 \\
  & &\textbf{{\footnotesize (0.9)}} &{\footnotesize (3.4)} &{\footnotesize (3.8)} &{\footnotesize (1.0)} &{\footnotesize (1.2)} &{\footnotesize (2.8)} &{\footnotesize (3.9)} &{\footnotesize (0.8)} &{\footnotesize (0.2)} \\
  &\multirow{2}{*}{Real} &51.2 &46.8 &45.5 &50.7 &43.9 &47.7 &46.2 &55.2 &\textbf{56.5} \\
  & &{\footnotesize (1.0)} &{\footnotesize (2.3)} &{\footnotesize (0.6)} &{\footnotesize (0.3)} &{\footnotesize (0.7)} &{\footnotesize (0.9)} &{\footnotesize (2.8)} &{\footnotesize (2.6)} &\textbf{{\footnotesize (0.4)}} \\
  &\multirow{2}{*}{Sketch} &33.5 &35.7 &37.5 &36.5 &28.9 &38.0 &37.4 &38.2 &\textbf{51.9} \\
  & &{\footnotesize (1.1)} &{\footnotesize (0.9)} &{\footnotesize (2.3)} &{\footnotesize (1.8)} &{\footnotesize (1.3)} &{\footnotesize (1.9)} &{\footnotesize (2.0)} &{\footnotesize (6.7)} &\textbf{{\footnotesize (1.9)}} \\
  &\cellcolor[HTML]{efefef} &\cellcolor[HTML]{efefef}42.9 &\cellcolor[HTML]{efefef}40.0 &\cellcolor[HTML]{efefef}40.2 &\cellcolor[HTML]{efefef}41.1 &\cellcolor[HTML]{efefef}36.7 &\cellcolor[HTML]{efefef}40.8 &\cellcolor[HTML]{efefef}40.3 &\cellcolor[HTML]{efefef}46.8 &\cellcolor[HTML]{efefef}\textbf{51.1} \\
  &\multirow{-2}{*}{\cellcolor[HTML]{efefef}Average} &\cellcolor[HTML]{efefef}{\footnotesize (0.5)} &\cellcolor[HTML]{efefef}{\footnotesize (1.5)} &\cellcolor[HTML]{efefef}{\footnotesize (0.5)} &\cellcolor[HTML]{efefef}{\footnotesize (0.0)} &\cellcolor[HTML]{efefef}{\footnotesize (0.3)} &\cellcolor[HTML]{efefef}{\footnotesize (0.4)} &\cellcolor[HTML]{efefef}{\footnotesize (0.3)} &\cellcolor[HTML]{efefef}{\footnotesize (1.5)} &\cellcolor[HTML]{efefef}\textbf{{\footnotesize (0.2)}} \\
  \bottomrule
  \end{tabular}
  \footnotesize{$^a$+GN means FedAvg+GN, $^b$+LN means FedAvg+LN}
\end{table}
\endgroup

\end{document}